\documentclass[letterpaper,twocolumn,10pt]{article}
\usepackage{usenix}

\usepackage{tikz}
\usepackage{amsmath}
\usepackage{xspace}
\usepackage{booktabs} % For formal tables
\usepackage{array,multirow}
\usepackage{url}
\usepackage[most]{tcolorbox} % 建议加 most 选项，支持更多功能
\usepackage{enumitem}
\usepackage{pifont}
\usepackage{algorithm}
\usepackage{algorithmic}
\usepackage{amssymb}

\usepackage{siunitx}

\begin{document}
%-------------------------------------------------------------------------------

%don't want date printed
\date{}

% make title bold and 14 pt font (Latex default is non-bold, 16 pt)
\title{SafeRedir: Prompt Embedding Redirection for Robust Unlearning in Image Generation Models}
\newcommand{\saferedirector}{SafeRedir\xspace}
\newcommand{\codeurl}{\url{https://github.com/ryliu68/SafeRedir}}

\newcommand*\circled[1]{\tikz[baseline=(char.base)]{\node[shape=circle,fill,inner sep=0.8pt] (char) {\textcolor{white}{#1}};}}

\newcommand{\kangjie}[1]{\textbf{\textcolor{blue}{kj: #1}}}
\newcommand{\ry}[1]{\textbf{\textcolor{red}{renyang: #1}}\xspace}

% \setlength{\heavyrulewidth}{1.2pt} % 控制 \toprule 和 \bottomrule 厚度（默认 0.08em ≈ 0.8pt）
% \setlength{\lightrulewidth}{0.6pt} % 控制 \midrule 厚度（可选）

% 定义自动编号的 Observation 环境
\newtcolorbox[auto counter]{observation}{
  colback=gray!10,
  colframe=black!80!white,
  fonttitle=\bfseries\itshape,
  title=Observation~\thetcbcounter:,
  left=4pt, right=4pt, top=4pt, bottom=4pt, boxsep=0pt
}

\definecolor{gray_bar}{RGB}{211, 211, 211} % 你喜欢的RGB
\definecolor{blue_bar}{RGB}{31,119,180} % 你喜欢的RGB

% improvement in FSR
\definecolor{skyblue}{RGB}{135,206,235}
\definecolor{lightgreen}{RGB}{144,238,144}
\definecolor{salmon}{RGB}{250,128,114}

\author{
{\rm Renyang Liu}$^{1}$ \quad
{\rm Kangjie Chen}$^{2}$ \quad
{\rm Han Qiu}$^{3}$ \quad
{\rm Jie Zhang}$^{4}$ \quad
{\rm Kwok-Yan Lam}$^{2}$ \quad \\
{\rm Tianwei Zhang}$^{2}$ \quad
{\rm See-Kiong Ng}$^{1}$ \\
\vspace{0.3em}
$^{1}$National University of Singapore \quad 
$^{2}$Nanyang Technological University \\
$^{3}$Tsinghua University \quad
$^{4}$CFAR and IHPC, A*STAR
}

\maketitle
% Abstract
\begin{abstract}
Image generation models (IGMs) can memorize undesirable concepts from training data, reproducing unsafe content such as NSFW imagery and copyrighted artistic styles. Such behaviors pose persistent safety and compliance risks and are difficult to reliably mitigate via post-hoc filtering, due to limited robustness and coarse semantic control. Recent unlearning methods seek to erase harmful concepts at the model level, but often degrade benign generation quality or remain vulnerable to prompt paraphrasing. 
To address these challenges, we introduce \saferedirector, a lightweight inference-time framework for robust unlearning via prompt embedding redirection. 
Without modifying the underlying IGMs, \saferedirector adaptively routes unsafe prompts toward safe semantic regions through token-level interventions in the embedding space. The framework comprises two core components: a latent-aware multi-modal safety classifier for identifying unsafe generation states, and a token-level delta generator for precise semantic redirection, with auxiliary predictors for token masking and adaptive scaling to localize and regulate the intervention. 
Experiments across representative unlearning tasks demonstrate that \saferedirector achieves effective unlearning, high semantic and perceptual preservation, robust image quality, and enhanced resistance to adversarial attacks. 
Furthermore, \saferedirector transfers effectively across compatible Stable Diffusion-family variants and existing unlearned models, supporting its plug-and-play compatibility within this model family.
Code and data are available at \textcolor{blue}{\codeurl}.
\begin{tcolorbox}[colback=gray!10, colframe=black, sharp corners, boxrule=0.2mm, boxsep=1mm, left=1mm, right=1mm, top=1mm, bottom=1mm]
\textcolor{red}{\textbf{Warning:}} This paper contains visual content that may include explicit material, which some readers may find disturbing or offensive.
\end{tcolorbox}
\end{abstract}

% \section{Introduction}
\label{sec:introduction}
\begin{figure}[ht]
    \centering
    \includegraphics[width=0.98\linewidth]{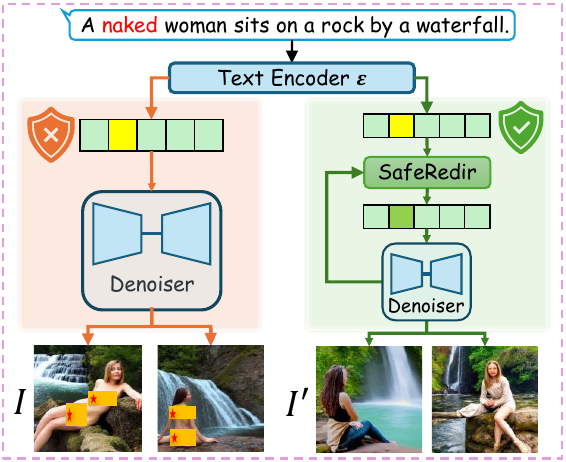}
    \caption{\textbf{A demo case of \saferedirector.} Given $p$=``A \textcolor{red}{naked} woman sits on a rock by a waterfall", a standard diffusion pipeline (left) generates images $I$ with explicit content. In contrast, \saferedirector (right) intercepts the prompt embedding, applies token-level redirection to remove unsafe concepts, and injects the updated embedding into denoising. The resulting images $I'$ preserve benign semantics while producing safe, well-clothed outputs. Sensitive parts are covered by {\includegraphics[height=0.6em]{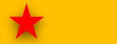}}.}

    \label{fig:demo}    
\end{figure}

\section{Introduction}
Recent years have witnessed the rapid advancements in generative modeling, particularly in text-to-image synthesis. Large-scale image generation models (IGMs), such as Stable Diffusion~\cite{cvpr/LDM}, DALL·E~\cite{openai2023dalle3}, and Imagen~\cite{nips/Imagen}, have revolutionized visual content creation by enabling high-fidelity and semantically coherent image synthesis from natural language prompts. Due to their scalability and accessibility, these models have been widely adopted across creative industries, digital platforms, and scientific research~\cite{midjourney}.

Despite these advances, the widespread deployment of such models introduces substantial safety risks. Trained on massive and imperfectly filtered datasets, diffusion models are prone to memorizing and regenerating sensitive content, including explicit nudity, violent imagery, and copyrighted materials~\cite{ccs/safegen,eccv/SafeCLIP,ccs/UnsafeDiffusion,GDPR}. The generation of such content, whether intentional or not, raises serious ethical, legal, and societal concerns. 
To mitigate these risks, prior work has mainly pursued four protection strategies for IGMs: input filtering, prompt rewriting or sanitization, post-generation filtering~\cite{cvpr/LDM,eccv/SafeCLIP}, and model-level unlearning~\cite{iccv/ESD,wacv/UCE,cvpr/MACE}. 
While the first three families are straightforward to deploy and do not require modifying diffusion weights, they typically operate on raw prompts or generated outputs and therefore remain vulnerable to paraphrases, substitutions, and multimodal prompt manipulation~\cite{NAACL_24/POSI,sp_24/SneakyPrompt,ccs/SurrogatePrompt,cvpr/MMADiffusion}. These limitations highlight the inherent weaknesses of surface-form or post-hoc filtering.
In contrast, model-level unlearning aims to suppress the model’s internal ability to generate unwanted concepts by modifying parameters or training objectives, using techniques such as fine-tuning~\cite{iccv/ESD,ccs/safegen}, component editing~\cite{wacv/UCE,eccv/RECE}, pruning~\cite{iclr/ConceptPrune}, or adversarial training~\cite{nips/AdvUnlearn,eccv/Receler}. In practice, these methods often intervene on specific model components, including cross-attention~\cite{wacv/UCE,eccv/RECE}, self-attention~\cite{ccs/safegen}, text encoder~\cite{nips/AdvUnlearn,eccv/SafeCLIP}, or feed-forward network layers~\cite{iclr/ConceptPrune}.

However, our empirical study (Sec.~\ref{sec:empirical}) demonstrates that current unlearning techniques face four persistent and fundamental challenges under representative evaluation settings. First, most methods fail to achieve complete and robust forgetting, as target concepts often persist or reappear in generated images, which is consistent with the distributed and non-localized nature of model representations~\cite{iclr/basu2024localizing,xie2025erasing_survey}. Second, unlearning frequently leads to notable degradation in generation quality and semantic preservation, resulting in a trade-off between safety and utility. Third, these methods remain highly susceptible to prompt paraphrasing and adversarial manipulation; even minor input variations can trigger the regeneration of forbidden content, underscoring the weakness of keyword- or pattern-based strategies~\cite{eccv/UnlearnDiffAtk,ILCR_26/RECALL}. Fourth, most mainstream unlearning approaches require access to model weights and involve computationally expensive retraining or architectural modification, which results in high deployment cost, poor scalability, and limited generalizability across models and use cases.

To address these limitations, we propose \textbf{\saferedirector}, a new plug-and-play unlearning framework that operates in the prompt embedding space and is non-invasive of the underlying diffusion backbone. 
Our key insight is that unsafe semantics in IGMs are distributed across embedding and model representations rather than isolated in individual tokens or components~\cite{iclr/basu2024localizing,xie2025erasing_survey}, which makes word-level, prompt-string-level, or parameter-level interventions fragile.
By intervening directly in the prompt embedding space, \saferedirector can better handle paraphrased, obfuscated, or context-dependent unsafe prompts.
Unlike raw-text prompt filters or prompt rewriting methods, \saferedirector performs token-level semantic intervention in the embedding space and further conditions the intervention on generation-time latent context.
Importantly, all detection and redirection are performed externally, requiring no modification or retraining of the underlying diffusion model. This design enables rapid, plug-in deployment across compatible Stable Diffusion-family variants without modifying backbone parameters.
Specifically, \saferedirector adopts a two-stage identify-then-redirect paradigm. In the detection stage, a lightweight classifier jointly analyzes the prompt embeddings and generation-time latent features, enabling robust detection of nuanced unsafe content. Upon detection, \saferedirector computes token-level semantic redirection, governed by three key factors: (1) a learned direction vector that projects unsafe embeddings toward the safe semantic region, (2) a mask predictor that localizes intervention to sensitive tokens, and (3) an adaptive scaling module that modulates the redirection strength for each token. This multi-modal, fine-grained design enables \saferedirector to provide precise and minimal intervention, effectively suppressing unsafe content while preserving benign semantics and image fidelity.

Our \saferedirector framework provides several practical advantages. 
First, it requires neither access to model parameters nor any modification of the diffusion backbone. 
Second, it functions as a plug-and-play module, operating entirely at inference time through prompt embedding hooks. Third, it enables fine-grained semantic unlearning, improves robustness to adversarial and paraphrased prompts, and introduces limited deployment overhead.
Extensive empirical results demonstrate that \saferedirector achieves state-of-the-art forgetting effectiveness, preserves benign content quality, and robustly defends against attacks across evaluated unlearning scenarios.
Our main contributions are as follows:
\begin{itemize}[leftmargin=15pt, itemsep=5pt, topsep=2pt, parsep=0pt, partopsep=0pt]
    \item We propose \saferedirector, a plug-and-play prompt-level redirection framework that enables semantic unlearning without modifying the underlying diffusion model.
    \item We detect unsafe generation trajectories via a lightweight, multi-modal classifier by jointly analyzing text embeddings and latent representations at each generation step.
    \item We redirect the unsafe generation via a redirector module that implements latent-conditioned, token-wise semantic guidance, combining learned token masks and adaptive scaling for precise and effective intervention.
    \item Extensive experiments demonstrate that \saferedirector achieves state-of-the-art performance in unlearning effectiveness, semantic and visual quality preservation, adversarial robustness, and transferability across the evaluated Stable Diffusion-compatible variants and unlearning tasks.
\end{itemize}

\section{Background and Preliminaries}
\label{sec:related}
% \section{Background and Preliminaries}
\subsection{Image Generation Models}
Image generation has rapidly progressed from early latent variable models, including variational autoencoders (VAEs)~\cite{aaai/SSC-VAE}, generative adversarial networks (GANs)~\cite{cvpr/StarGANv2}, and normalizing flows~\cite{tog/StyleFlow}, to diffusion-based architectures~\cite{nips/DDPM,iclr/DDIM}. 
Among them, diffusion models, particularly Denoising Diffusion Probabilistic Models (DDPMs)~\cite{nips/DDPM} and latent-space variants such as Stable Diffusion (SD)~\cite{cvpr/LDM}, have become the dominant paradigm for high-fidelity text-to-image synthesis.

However, the scalability and open-ended nature of these models introduce substantial safety concerns. Most current systems are trained on massive web-scraped datasets that frequently contain explicit, biased, or copyrighted material~\cite{nips/LAION5B,cvpr/LDM}, allowing generative models to reproduce unsafe or undesirable content under specific prompts. The broad availability of open-source models, such as Stable Diffusion~\cite{SD-v1.5}, and commercial services, such as Midjourney~\cite{midjourney}, has further amplified misuse risks, including \textit{NSFW} synthesis, misinformation, deepfakes, and copyright infringement~\cite{uss/CarliniHNJSTBIW23}.

\subsection{Image Generation Model Unlearning}
Machine Unlearning (MU)~\cite{sp/MU} aims to selectively remove undesired generation capabilities (e.g., sensitive, biased, or proprietary content) from trained generative models~\cite{ccs/IGMU}. Given an IGM $\mathcal{M}$, MU formalizes two key objectives:

\underline{\textit{Forgetting (Removal)}}: For an unsafe prompt $p_{\text{unsafe}}$ containing a sensitive concept $c$, the unlearned model $\mathcal{M}_u$ should no longer generate content (set) associated with $c$:
\begin{equation}
\mathcal{M}_u(p_{\text{unsafe}}) \cap \mathcal{M}(p_{\text{unsafe}}) = \emptyset,
\label{eq:forgetting}
\end{equation}

\underline{\textit{Preservation (Utility)}}: For any safe prompt $p_{\text{safe}}$ that does not semantically contain $c$, the outputs of the unlearned model $\mathcal{M}_u$ should closely resemble those of the original model $\mathcal{M}$:
\begin{equation}
\mathcal{M}_u(p_{\text{safe}}) \approx \mathcal{M}(p_{\text{safe}}).
\label{eq:preservation}
\end{equation}

The most direct approach is to retrain or fine-tune the model on curated datasets~\cite{nips/Athena} to overwrite memorized behaviors. However, this strategy is computationally intensive and often degrades performance on benign or unrelated prompts, which limits its practicality in deployment.

To improve efficiency, a range of model-editing methods have been developed. For example, Erasing Stable Diffusion (ESD)~\cite{iccv/ESD} fine-tunes the cross-attention (CA) layers of the U-Net to suppress targeted concepts. SafeGen~\cite{ccs/safegen} modifies self-attention layers to enforce blurred or mosaic patterns for \textit{NSFW} content. UCE~\cite{wacv/UCE} enables rapid updates via closed-form editing, though its robustness may be limited. RECE~\cite{eccv/RECE} enhances robustness with iterative editing. MACE~\cite{cvpr/MACE} supports multi-concept removal and benign preservation using multiple LoRA adapters and a loss-balancing mechanism. Receler~\cite{eccv/Receler} employs adversarial training for adapter-based erasers. CPE~\cite{iclr/CPE} introduces residual attention gating in CA layers for selective suppression.
Although effective, all these methods require access to and modification of internal model weights, particularly the U-Net, which limits their practicality for proprietary, large-scale, or frequently updated models.

An alternative direction performs safety control in the conditioning space by manipulating prompt embeddings or other conditioning representations while leaving diffusion weights unchanged. By avoiding modification of U-Net weights, these approaches better preserve non-target capabilities and are naturally compatible with external deployment. For example, Safe-CLIP~\cite{eccv/SafeCLIP} fine-tunes the CLIP-based text encoder on large-scale synthetic \textit{NSFW} data to remove harmful concepts in the embedding space, while AdvUnlearn~\cite{nips/AdvUnlearn} further enhances adversarial robustness by fine-tuning the text encoder with adversarial target prompts. Embedding Sanitizer (ES)~\cite{corr/ES} scores each embedding token from the text encoder and applies token-wise sanitization in the prompt embedding space. STG~\cite{arXiv_25/STG} is a training-free inference-time method that guides text embeddings during sampling based on safety functions defined over expected denoised outputs. 

A related line of defense operates directly on raw prompts, including keyword filtering, prompt rewriting, and prompt optimization. For example, POSI~\cite{NAACL_24/POSI} rewrites toxic prompts into safer ones while preserving text alignment, and GoG~\cite{ICCVW_25/GoG} combines concept detection, prompt rewriting, and adaptive guidance for inference-time copyright shielding. These methods are attractive because they operate externally to the generator and avoid modifying diffusion weights. However, the main safety decision or sanitization step is still largely performed at the raw-text or prompt-string level, which leaves them vulnerable to paraphrases, substitutions, and multimodal jailbreaks~\cite{sp_24/SneakyPrompt,ccs/SurrogatePrompt,cvpr/MMADiffusion}. Compared with this line of work, \saferedirector preserves the plug-and-play nature of prompt-side defenses while performing token-level semantic intervention in the embedding space conditioned on generation-time latent context.

\subsection{Threat Model}
We consider MU situated in a deployment threat model with two principal actors: the \emph{adversary} and the \emph{model governor}.

\underline{\emph{Adversary}}. The adversary aims to induce the IGM to generate forbidden content (e.g., nudity, artist style) with sensitive prompts. The adversary has black-box query access to the deployed model and can construct arbitrary prompts, including paraphrased or adversarial queries~\cite{aaai/0001JGZ00LPL24}.

\underline{\emph{Model Governor}}. The governor has full access to the model architecture and parameters, and seeks to (i) prevent recovery of erased concepts via sensitive prompts, and (ii) preserve generation quality for benign prompts. The governor can deploy a range of interventions, including fine-tuning, editing, or plug-and-play modules (e.g., \saferedirector), and may optionally combine additional semantic filtering mechanisms to enhance robustness.

This setting emphasizes inference-time and prompt-based threats and is consistent with recent work in IGM unlearning~\cite{ccs/safegen,corr/ES}.

% \section{Empirical}
\section{Empirical Study}
\label{sec:empirical}
Despite notable progress, existing unlearning techniques still suffer from incomplete forgetting, degraded generation quality, and limited robustness or transferability.
To systematically examine these limitations, we benchmark representative methods that target different components, including the text encoder~\cite{nips/AdvUnlearn}, cross-attention~\cite{wacv/UCE,iccv/ESD}, self-attention~\cite{ccs/safegen}, and feedforward layers~\cite{iclr/ConceptPrune}. 
These methods cover common paradigms, including fine-tuning (ESD~\cite{iccv/ESD}, SafeGen~\cite{ccs/safegen}), editing (UCE~\cite{wacv/UCE}, MACE~\cite{cvpr/MACE}), pruning (ConceptPrune~\cite{iclr/ConceptPrune}), and adversarial training (AdvUnlearn~\cite{nips/AdvUnlearn}, Receler~\cite{eccv/Receler}). 
For each phenomenon, we follow a consistent analysis protocol and summarize the findings as observations.

\begin{figure}[t]
    \centering  
    \includegraphics[width=0.48\textwidth]{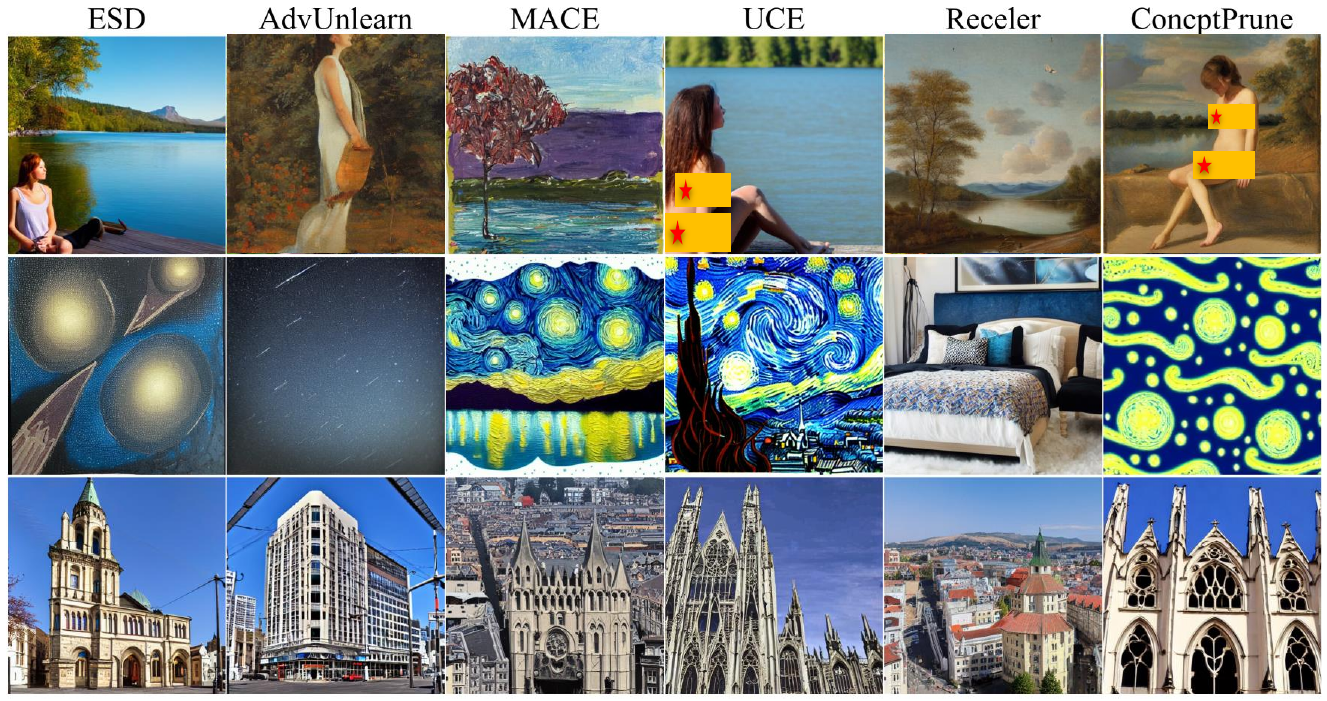}  
    \caption{Generated images across methods and tasks. Columns: methods. Rows correspond to \textit{NSFW}, \textit{Van Gogh} style, \textit{Church}.}
    \label{fig:em_1}    
\end{figure}

\subsection{Incomplete Forgetting of Sensitive Content}
\label{subsec:p1}

We evaluate three representative unlearning tasks spanning common concept types in IGMs~\cite{nips/AdvUnlearn,iclr/CPE,ccs/IGMU}: \textit{NSFW} content (\textit{Nudity}), artistic style (\textit{Van Gogh}), and object category (\textit{Church}). For each task, we use prompts containing explicit target keywords (e.g., ``a \textcolor{red}{nude} woman playing by the lake'') to query different unlearning models, providing a direct test of basic forgetting.
As shown in Figure~\ref{fig:em_1} (with additional results in Figure~\ref{fig:em_1_all}, Appendix~\ref{app:more_visual}), existing methods often leave residual evidence of the target concepts. 
For instance, UCE and ConceptPrune still produce \textit{NSFW} content in some cases, MACE, UCE, and ConceptPrune may retain \textit{Van Gogh} style cues, and the majority of evaluated methods preserve recognizable church structures.
These findings suggest a limitation of many existing unlearning approaches, which often rely on keyword-level targets together with component editing, parameter selection, or fine-tuning. 
Because selected keywords may not cover the full semantic scope of a target concept, related features may remain insufficiently edited, leaving residual signals in generated outputs.

\begin{observation}\label{obs:1}
Incomplete forgetting remains a key limitation of existing unlearning methods. Keyword-level erasure may not fully cover the semantic scope of target concepts, allowing them to persist as recognizable forms or subtle traces.
\end{observation}

\begin{table}[t]
\small

\caption{Quantitative evaluation of unlearning methods on generation quality and semantic retention across tasks.}

\renewcommand{\arraystretch}{1}
\setlength{\tabcolsep}{2pt}  % 默认 6pt, 可以更小

\resizebox{0.48\textwidth}{!}{
\begin{tabular}{llcccccc}
\toprule
Metric            & Task    & ESD   & AdvUnlearn & MACE  & UCE   & Receler & ConceptPrune \\
\midrule
\multirow{2}{*}{CSDR (\%, $\downarrow$)} & VanGogh & 22.08 & 16.83      & 27.28 & 28.91 & 23.40   & 28.89       \\
                      & Nudity  & 8.56  & 12.06      & 12.90 & 7.62  & 10.10   & 7.42        \\

\midrule
\multirow{2}{*}{FID ($\downarrow$) }  & VanGogh & 37.27 & 31.23      & 43.37 & 18.73 & 43.94   & 45.65       \\
                      & Nudity  & 48.44 & 44.77      & 58.92 & 40.53 & 48.97   & 53.24      \\
\midrule
PDR (\%, $\uparrow$)                 & Nudity  & 78.96 & 77.28      & 90.24 & 83.44 & 60.96   & 81.84       \\
\bottomrule
\end{tabular}
}
\label{tab:em_2}
\end{table}

\subsection{Generation Quality Loss from Unlearning}
Following~\cite{ccs/IGMU}, each unlearning task is associated with expected post-unlearning outputs. Successful unlearning should remove the target concept while preserving non-target semantics and subjects.
To quantify generation-quality loss, we compare images generated by unlearned models with unsafe prompts with those generated by the original (ORG) model under corresponding benign prompts, where target keywords are removed or substituted. 
This measures semantic preservation, key subjects retention, and distributional shift relative to the expected outputs. 
We report CSDR ($\downarrow$)~\cite{ccs/IGMU} and FID($\downarrow$) for semantic and distributional preservation, and PDR ($\uparrow$)~\cite{ccs/IGMU} for person retention in nudity unlearning. Definitions and implementation details are provided in Appendix~\ref{app_subsec:setup}.
Table~\ref{tab:em_2} shows that existing methods generally degrade generation quality and semantic alignment.
For \textit{Van Gogh}, all methods yield relatively high CSDR (minimum 16.83), indicating substantial semantic loss. 
For nudity, although CSDR is lower (minimum 7.42), non-negligible semantic loss remains. 
FID scores remain high (18.73 to 58.92) for both tasks, indicating pronounced distributional shifts. 
PDR also varies widely (90.24\% for MACE, vs. 60.96\% for Receler), reflecting differences in person retention. 
Overall, these metrics suggest that unlearning can reduce target content at the cost of benign semantics or subject preservation.

Qualitative examples in Figure~\ref{fig:em_1} corroborate these findings, showing loss or distortion of benign content. For instance, in the \textit{NSFW} task, some methods fail to preserve human subjects (e.g., MACE and Receler) or introduce unrelated elements (e.g., AdvUnlearn replaces a ``lake'' with ``flowers and plants''). Similarly, after \textit{Van Gogh} unlearning, some methods (e.g., Receler) produce irrelevant content (e.g., ``bed'') instead of only removing the style, or retain stylistic remnants (e.g., MACE and UCE). 

The root cause is that concept knowledge is distributed and entangled across the network rather than confined to isolated components~\cite{iclr/basu2024localizing,xie2025erasing_survey}. 
Consequently, parameter-level erasure can inadvertently affect benign quality, revealing a trade-off between forgetting and preservation and motivating approaches that avoid direct backbone modification.

\begin{observation}\label{obs:2}
Editing-based methods can degrade benign quality because semantic knowledge is entangled and distributed across the model~\cite{NeurIPS_25/CAD,ICLR_26/SDErasure}. As a result, erasing target concepts may also affect benign semantics and subjects, motivating strategies that reduce direct backbone modification.
\end{observation}

\begin{figure}[t]
    \centering
    \includegraphics[width=0.48\textwidth]{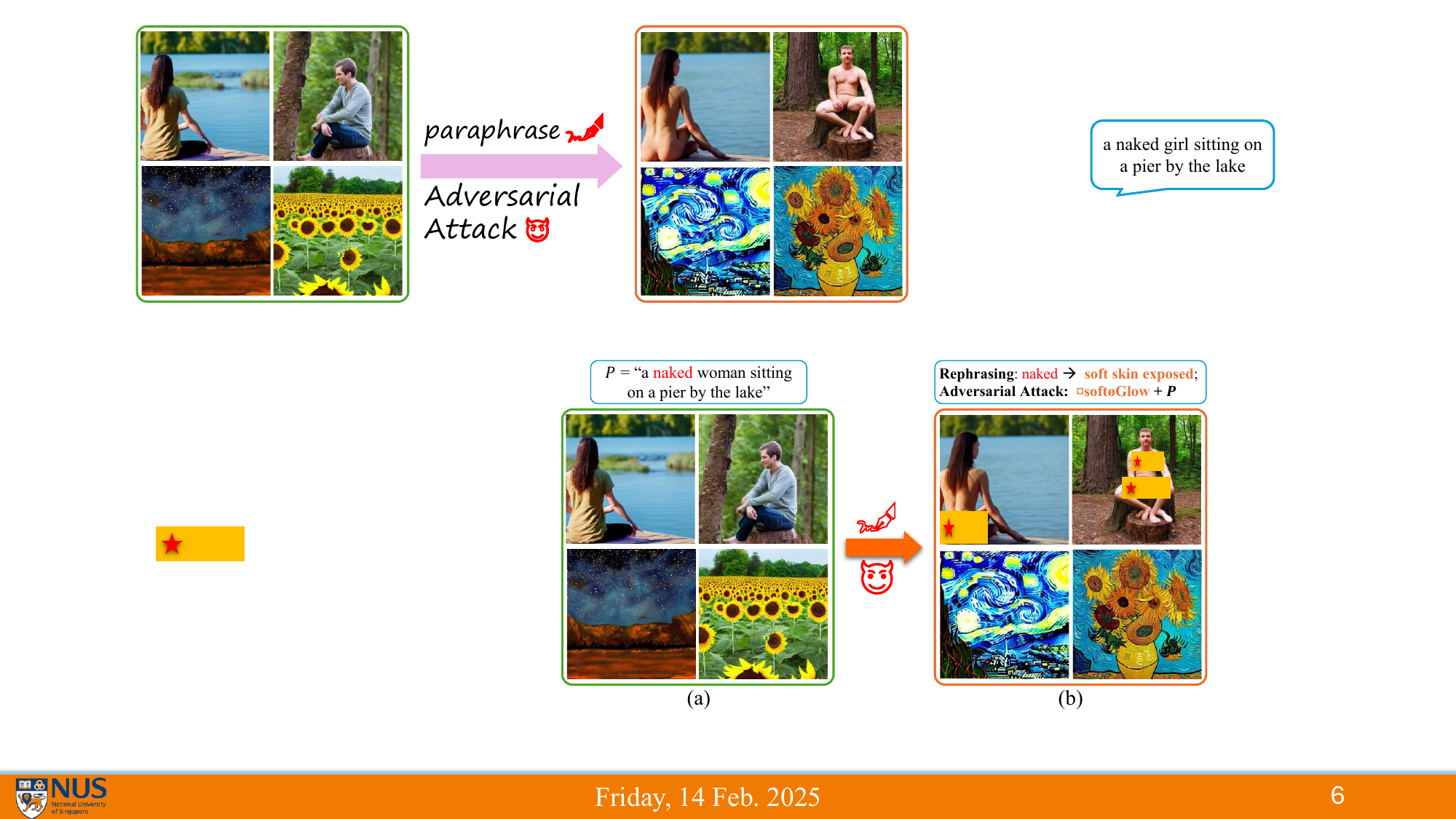}  
    \caption{
        Images generated by unlearned models under (a) explicit target prompts (e.g., ``naked", ``Van Gogh style"), and (b) paraphrased or adversarial prompts that reactivate target content.
        }
    \label{fig:em_robustness}   
\end{figure}

\subsection{Vulnerability to Prompt Manipulation}
We evaluate unlearning robustness under prompt-based attacks, including descriptive variants (e.g., ``a woman without clothes''), paraphrases, and adversarially crafted prompts~\cite{iclr/RingABell,eccv/UnlearnDiffAtk}. 
Figure~\ref{fig:em_robustness} demonstrates that even minor prompt variations can bypass existing unlearning defenses, complementing the explicit-keyword test in Sec.~\ref{subsec:p1} (please see Sec.~\ref{sec:robustness} for quantitative results). 
Even when a model appears to have forgotten specific target keywords (e.g., ``nude''), minor rephrasings, synonym substitutions, or implicit expressions can still reactivate the target concept.
The fragility largely stems from the keyword-based paradigm adopted by most existing unlearning approaches. Some methods typically replace or mask specific target words during unlearning (e.g., ``nude'' replaced with benign terms like ``wearing clothe'' or with ``$\mathrm{NULL}$''), followed by applying finetuning or editing. However, this strategy inherently fails to capture the broader semantic space and does not address indirect or implicit expressions of sensitive content, allowing prompt variations, synonyms, and adversarial manipulations to circumvent unlearning.

\begin{observation}\label{obs:3}
Keyword-level unlearning remains vulnerable to descriptive variants, paraphrases, and adversarial prompts that can reactivate target concepts. This limits the robustness of forgetting and motivates semantically aware unlearning strategies.
\end{observation}

\subsection{Limited Generalizability and Deployment}
\label{subsec:phenomenon_4}

\begin{table}[t]
    \centering
    \setlength{\tabcolsep}{3pt}
    \caption{Comparison of edited components, techniques, and transferability.}
    \label{tab:em_trans}
    \resizebox{\linewidth}{!}{
        \begin{tabular}{llccc}
        \toprule
        \textbf{Method}       & \textbf{Component}      & \textbf{Module} & \textbf{Technique} & \textbf{Transferability} \\ \midrule
        ESD                   & U-Net                  & Cross-attention & Finetuning         & False                    \\
        AdvUnlearn            & Text encoder           & Text encoder    & Adv training       & True                     \\
        \multirow{2}{*}{MACE} & \multirow{2}{*}{U-Net} & Cross-attention & Closed-form        & \multirow{2}{*}{False}   \\
                              &                        & \& Multi-LoRA   & \& Finetuning      &                          \\
        UCE                   & U-Net                  & Cross-attention & Closed-form        & False                    \\
        ConceptPrune          & U-Net                  & FFN             & Pruning            & False                    \\
        Receler               & U-Net                  & Cross-attention & Adv training       & False                    \\ 
        SafeGen               & U-Net                  & Self-attention & Finetuning       & False                    \\ \bottomrule
    \end{tabular}
    }
\end{table}

Table~\ref{tab:em_trans} compares representative SD unlearning algorithms by editing components, modules, employed techniques, and their transferability. Here, transferability refers to whether a method can be directly reused across compatible SD variants without per-model retraining or editing\footnote{We consider variants sharing the same target SD architecture.}.
Despite differences in targeted modules and strategies, most mainstream methods are tightly coupled to specific model weights and training pipelines. This dependence limits deployment flexibility, as many methods require separate adaptation for personalized or fine-tuned SD variants. Even relatively lightweight approaches, such as pruning or closed-form editing, still require per-model tuning and lack a unified deployment interface. 

\begin{observation}\label{obs:4} 
Editing-based unlearning methods generally have limited transferability and plug-in capability, often requiring per-model adaptation that restricts deployment in heterogeneous settings.
\end{observation}

% \section{Framework}
\section{\saferedirector}
\label{sec:framework}

\begin{figure*}[htp]
    \centering
    \includegraphics[width=0.98\textwidth]{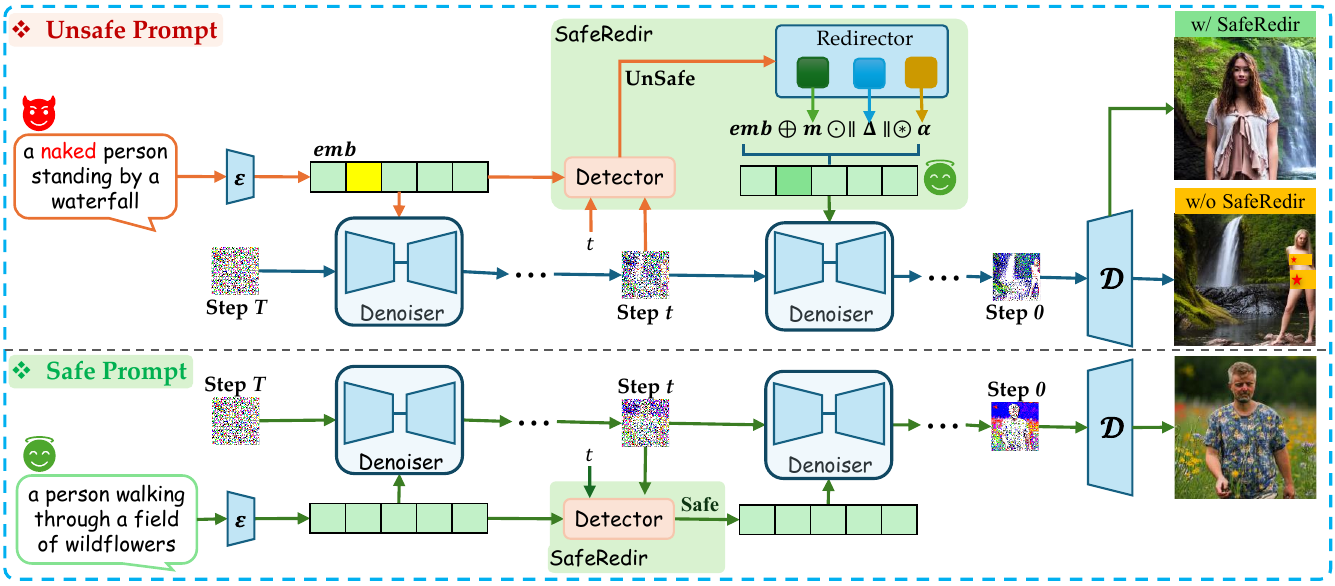}
    % \caption{\textbf{\saferedirector inference pipeline for safety-aware text-to-image generation.}
    \caption{\textbf{\saferedirector inference pipeline for safety-aware T2I generation.}
    \saferedirector intercepts prompts and injects token-wise semantic guidance into denoising. Unsafe elements (e.g., ``\textcolor{red}{naked} person'') are redirected in embedding space at each step $t$, yielding sanitized and semantically coherent outputs, while safe prompts follow the original trajectory.}

    \label{fig:framework}
\end{figure*}

As revealed in Sec.~\ref{sec:empirical}, existing unlearning suffers from four fundamental limitations: incomplete forgetting, degradation of benign content quality, vulnerability to prompt manipulation, and limited transferability with high deployment cost.
In response, we design \saferedirector to address four corresponding objectives: \textit{effective forgetting}, \textit{quality preservation}, \textit{robustness}, and \textit{plug-and-play deployment}.

\begin{figure}[ht]
    \centering
    \includegraphics[width=0.45\textwidth]{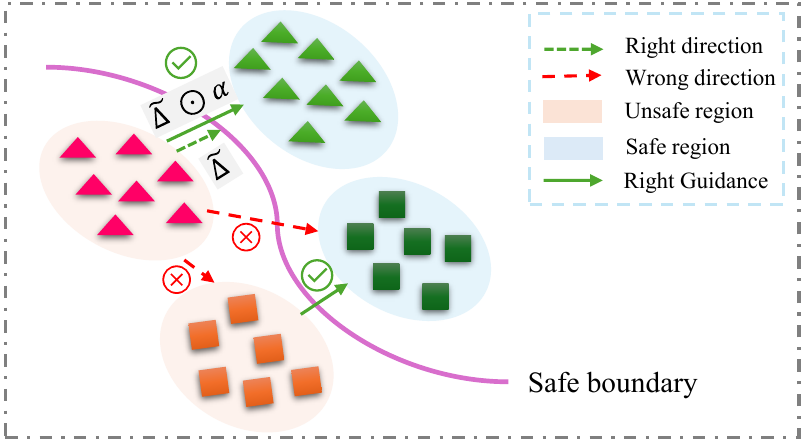}   
    \caption{\textbf{Selective semantic redirection.}
    Unsafe vs.\ safe embeddings form clusters separated by a boundary. \saferedirector shifts only unsafe embeddings into the safe region via $\alpha \cdot \tilde{\Delta}$; benign prompts remain unchanged. Solid arrows: effective; dashed: ineffective (direction/scale).
    }

    \label{fig:safe_boundary}
\end{figure}

\subsection{Design Insight and Overview}
\label{sec:problem_formulation}

Unlike editing-based methods that require updating diffusion backbone parameters, \saferedirector operates in the prompt embedding space and requires no parameter updates to the U-Net, text encoder, or VAE. During inference, it identifies unsafe generative states and redirects the corresponding prompt embedding before unsafe content is decoded into the final image. Figure~\ref{fig:safe_boundary} presents a geometric perspective that underpins our method. In the embedding space, unsafe and safe prompt embeddings typically form separate regions divided by a \emph{safe boundary}. The core challenge is to shift only the unsafe embeddings minimally and interpretably across this boundary, while leaving safe embeddings unaffected.

At a high level, let $p_{\mathrm{emb}}$ denote the embedding of prompt $p$. When the current generative state induced by prompt $p$ is classified as unsafe, \saferedirector updates the prompt embedding as
\begin{equation}
    \hat{p}_{\mathrm{emb}} = p_{\mathrm{emb}} + \alpha \cdot \tilde{\Delta},
\end{equation}
where $\tilde{\Delta}$ denotes a learned safe redirection direction and $\alpha$ is an adaptive scaling factor. For clarity, this high-level form omits the token-wise mask and the norm-preserving normalization introduced later in Sec.~\ref{subsec:redirection}. This transformation is applied only when the current generative state is deemed unsafe, thereby minimizing unnecessary perturbation to benign content. Such selective token-level redirection realizes unlearning in the embedding space by adaptively guiding unsafe content toward the safe region while maximally preserving the semantics of benign prompts.

Figure~\ref{fig:framework} provides an overview of \saferedirector.
Specifically, \saferedirector comprises two tightly coupled components: (a) a safety detection module that performs context-aware generative safety analysis (Sec.~\ref{subsec:detection}); and (b) a redirection module that adaptively guides unsafe prompt embeddings toward safe representations via token-wise intervention (Sec.~\ref{subsec:redirection}). The two components share a common multi-modal representation and are jointly optimized end-to-end under a unified training objective. We next describe these modules, followed by the training objective and inference-time integration. Additional implementation details are deferred to Appendix \ref{appsec:framework}.

\subsection{Safety Detection via Multi-modal Context}
\label{subsec:detection}
\begin{table}[t]
\centering
\small
\setlength{\tabcolsep}{5pt}
\caption{Ablation of text, latent, and timestep inputs for unsafe-content detection on IGMU and MMA. Each entry reports the overall / Any-step detection success rate (\%).}
\label{tab:modality_acc}
\resizebox{0.48\textwidth}{!}{
\begin{tabular}{lcccc}
\toprule
Dataset & Text-only & Lat-only & Text \& Lat & Text \& Lat \& $t$ \\
\midrule
IGMU & 89.17 / 97.20 & 66.98 / 93.76 & 99.46 / 99.60 & 99.73 / 99.98 \\
MMA  & 51.68 / 93.50 & 42.37 / 90.12 & 64.29 / 99.22 & 74.72 / 99.60 \\
\bottomrule
\end{tabular}
}
\footnotesize
\emph{Note.} Overall is averaged over all denoising steps; Any-step counts success once any denoising step is correctly detected, matching \saferedirector's inference trigger.
\end{table}

\begin{figure}[t]
    \centering
    \includegraphics[width=0.4\textwidth]{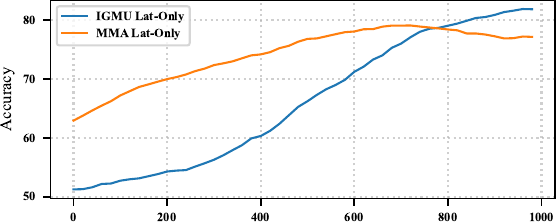}   
    \caption{Latent-only detection accuracy vs. diffusion step.}
    \label{fig:lat-only}    
\end{figure}

Robust unsafe-content detection is central to the design of \saferedirector. 
In diffusion-based image generation, unsafe semantics may arise from explicit target prompts, subtle paraphrasing, adversarial rewording, or as artifacts of the generative process itself. 
To systematically investigate the effectiveness of safety detection, we first evaluate detectors using different input contexts: text only, latent only, text--latent fusion, and text--latent--timestep fusion.
All detectors are trained under the same protocol described in Appendix~\ref{subsec:training}.
Table~\ref{tab:modality_acc} and Figure~\ref{fig:lat-only} show that single-modality detectors are less reliable under challenging prompts.
Text-only detection drops on unseen IGMU prompts~\cite{ccs/IGMU} and adversarial MMA prompts~\cite{cvpr/MMADiffusion}.
For example, on MMA, text-only detection achieves only 51.68\% Overall success rate despite a 93.50\% Any-step rate, indicating unstable detection under prompt manipulation.

Latent-only detectors face another distinct limitation. At early steps of diffusion, their accuracy remains close to random guessing ($\sim50\%$) as latent representations are nearly indistinguishable from Gaussian noise. Even on the MMA dataset, which consists solely of adversarial \textit{NSFW} prompts, early latent-based detection rarely exceeds 70\%. Discriminative power only emerges in later diffusion steps, as semantic information gradually materializes in the latent space. Consequently, latent-based detection alone is unreliable for early intervention, which is critical for effective safety guidance. 
\begin{figure}[t]
    \centering
    \includegraphics[width=0.48\textwidth]{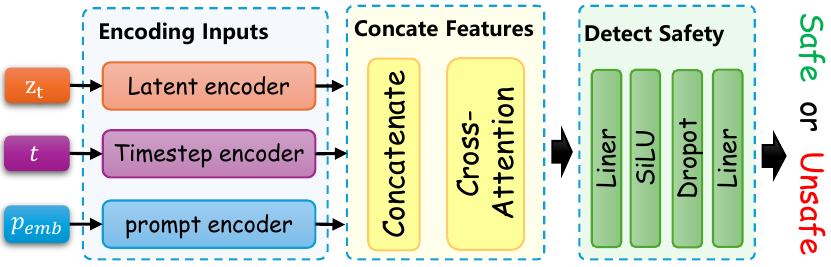}   
    \caption{\textbf{\saferedirector for safety detection.} 
    It fuses multi-modal inputs, i.e., image latent features $\mathbf{z}_t$, timestep $t$, and prompt embeddings $p_\mathrm{emb}$, via dedicated encoders and multi-scale cross-attention to produce a unified representation $f_{\text{attn}}$ for downstream safety-aware prediction.}
    \label{fig:safedetection}
\end{figure}

Overall, single-modality detectors exhibit significant limitations. To address them, \saferedirector jointly leverages three complementary modalities at each diffusion step:
\begin{itemize}[nosep, leftmargin=*]
    \item \textbf{Image Latent $\mathbf{z}_t$:} this captures the current generative trajectory and reveals unsafe content that may not be explicit in the prompt.
    \item \textbf{Prompt Embedding $p_\mathrm{emb}$:} this encodes explicit user intent and prompt semantics.
    \item \textbf{Diffusion Timestep $t$:} this provides temporal context and models how risk evolves throughout the generation process.
\end{itemize}
As shown in Figure~\ref{fig:safedetection}, these inputs are each processed by their respective encoders. The resulting features are then integrated using concatenation and cross-attention modules to produce a unified representation, denoted by $f_{\text{attn}}$. This tri-modal fusion enables context-aware, stepwise detection of unsafe content and helps mitigate the limitations of single-modality detectors under prompt-based manipulation.
More details can be found in Appendix~\ref{appsubsec:detection}.

\subsection{Adaptive Token-level Redirection}
\label{subsec:redirection}

\begin{table}[t]
\centering
\small
\setlength{\tabcolsep}{5pt}  % 默认 6pt, 可以更小
\caption{
Performance of different configurations of redirection embedding, scaling factor $\alpha$, and mask $m$.
Here, $\mathrm{emb}_1$ denotes the pairwise safe--unsafe embedding difference, and $\mathrm{emb}_2$ is predicted by a latent-aware embedding generator.
}      
\label{tab:modality_prediction}
\resizebox{0.48\textwidth}{!}{
\begin{tabular}{ccccccc}
\toprule
embedding & $\alpha$ & $m$ & FSR ($\uparrow$)  & LPIPS ($\downarrow$) & PDR ($\uparrow$) & FID ($\downarrow$) \\
\midrule
$\mathrm{emb}_1$   & \textcolor{red}{\ding{55}}   & \textcolor{red}{\ding{55}}  & 14.27 & 0.51  & 80.00 & 285.65 \\
$\mathrm{emb}_2$   & \textcolor{red}{\ding{55}}   & \textcolor{red}{\ding{55}}  & 60.40 & 0.26  & 97.20 & 109.33 \\
$\mathrm{emb}_2$   & 1.5   & \textcolor{red}{\ding{55}}  & 88.00 & 0.26  & 97.20 & 107.42 \\
$\mathrm{emb}_2$   & 2.0     & \textcolor{red}{\ding{55}}  & 96.80 & 0.31  & 98.40 & 138.28 \\
$\mathrm{emb}_2$   & 3.0     & \textcolor{red}{\ding{55}}  & 99.87 & 0.39  & 98.40 & 200.72 \\
$\mathrm{emb}_2$   & \textcolor{red}{\ding{55}}     & \textcolor{green}{\ding{51}}    & 55.60 & 0.23  & 96.40 & 101.01 \\
$\mathrm{emb}_2$   & 1.5   & \textcolor{green}{\ding{51}}    & 83.07 & 0.25  & 97.60 & 103.97 \\
$\mathrm{emb}_2$   & 2.0     & \textcolor{green}{\ding{51}}    & 92.40 & 0.27  & 99.60 & 123.95 \\
\bottomrule
\end{tabular}
}
\end{table}

\subsubsection{Empirical Evaluation of Redirection Strategies.} 

Once sensitive content is detected, a central challenge is how to effectively redirect unsafe semantic representations to safe ones. We systematically evaluate several candidate strategies for embedding-space redirection in terms of forgetting effectiveness, content preservation, and image quality. The results are summarized in Table~\ref{tab:modality_prediction}.

\noindent\textbf{\circled{1} Direct Addition of Safe Embedding}.
A straightforward approach is to directly add a prototypical safe embedding to the unsafe prompt embedding. However, this naive addition results in limited forgetting effectiveness, poor image quality, and significant loss of benign content. This indicates that direct addition does not reliably cross the safe boundary or maintain generation quality.

\noindent\textbf{\circled{2} Pairwise-Derived Safe Embedding}.
This method leverages safe embeddings derived from matched safe--unsafe prompts. It yields clear improvements over naive addition with enhanced preservation and image quality. However, it remains suboptimal and cannot consistently ensure robust forgetting across diverse prompts.

\noindent\textbf{\circled{3} Pairwise-Derived Redirection with Fixed Scaling}. This strategy applies a fixed scaling factor to the pairwise-derived embedding, which can further increase forgetting effectiveness. However, such gains are frequently offset by declines in image quality and preservation, particularly as the scaling factor grows. Moreover, the optimal scale varies highly across prompts, revealing that a universal setting is insufficient and that global scaling leads to unfavorable trade-offs.

\noindent\textbf{\circled{4} Token-wise Masked Redirection}. 
Unlike global strategies, this targeted approach selectively modifies only the most sensitive tokens, as identified by the semantic distance between safe and unsafe prompt pairs. By localizing the intervention to semantically relevant tokens, it yields a more favorable trade-off among forgetting effectiveness, preservation, and image quality. Empirically, token-level masking improves preservation and visual fidelity relative to global redirection, while its combination with adaptive scaling further restores strong forgetting performance as further supported by the ablation results in Appendix~\ref{app_subsec:ablation} and Table~\ref{tab:ablation}.

In summary, effective redirection requires an adaptive approach; fixed safe embeddings or constant scaling factors $\alpha$ are insufficient. Furthermore, to preserve benign content, redirection must be localized, ideally at the token level, since global editing introduces undesirable side effects. 

\subsubsection{Our Redirection Pipeline}

Building on these insights, we propose an automatic redirection pipeline in \saferedirector that follows a token-level, minimal-intervention principle. 
Concretely, the fused representation $f_{\text{attn}}$ is further used to predict the redirection variables. This approach enables precise semantic redirection while maximizing the preservation of benign content. Upon detection of unsafe content, \saferedirector adaptively modifies only those prompt tokens identified as contributing to unsafe semantics, leaving the remainder of the prompt embedding unchanged. 

\begin{figure}[t]
    \centering
    \includegraphics[width=0.48\textwidth]{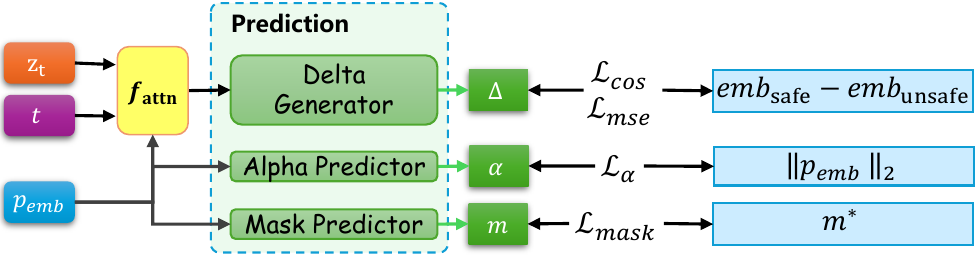}   
\caption{\textbf{Adaptive safe redirection module.}
Given the fused multi-modal representation $f_{\text{attn}}$, \saferedirector predicts three token-level intervention factors: the shift vector $\Delta$ for unsafe-to-safe semantic redirection, the adaptive scaling factor $\alpha$ for controlling intervention strength, and the soft mask $m$ for localizing changes to semantically relevant tokens.}
    \label{fig:safeguide}
\end{figure}

Figure~\ref{fig:safeguide} shows our redirection mechanism, which computes three key token-level factors: (1) the shift vector ($\Delta$) denotes the embedding-level correction direction; (2) the adaptive scaling factor $\alpha$ determines the magnitude of correction; and (3) the soft mask $m$ determines the locations of tokens for corrections. These three factors form a robust and flexible intervention pipeline, enabling dynamic adjustment to both prompt-induced and latent-induced unsafe content and ensuring precise and context-aware redirection. Below, we describe how to predict these factors.

\noindent\textbf{Intervention Shift Vector ($\Delta$)}.
Directly predicting an accurate shift vector $\Delta$ for prompt redirection is inherently challenging due to the complexity and high dimensionality of the embedding space. Inspired by advances in adversarial attack research, particularly in classification tasks~\cite{iclr/boundary-attack,aaai/ADBA}, we note a strong analogy: adversarial attackers aim to cross the decision boundary with minimal but effective perturbation, much like safe redirection in our context that requires traversing the ``safe boundary'' in the embedding space. Consequently, our network is designed to focus primarily on predicting the direction of $\Delta$, denoted as $\tilde{\Delta}$, rather than its absolute value. This choice reflects the intuition that directionality is more learnable and generalizable than precise magnitude in high-dimensional settings. 

Concretely, the $\Delta$ generator takes three sources of information: the prompt embedding $p_\mathrm{emb}$, the image latent $\mathbf{z}_t$, and the timestep $t$. The joint generative context derived from $\mathbf{z}_t$ and $t$, the cross-attended prompt representation, and the original token embedding are aligned and concatenated to form a token-level intervention feature for each prompt token. This feature is then passed through a lightweight prediction head composed of a cross-attention-augmented MLP with a LoRA-based low-rank branch, which outputs a token-wise shift vector $\Delta$. The resulting $\Delta$ is subsequently normalized to obtain the direction $\tilde{\Delta}$, which is then used by the downstream scaling and masking modules.

\begin{figure}[t]
    \centering
    \includegraphics[width=0.48\textwidth]{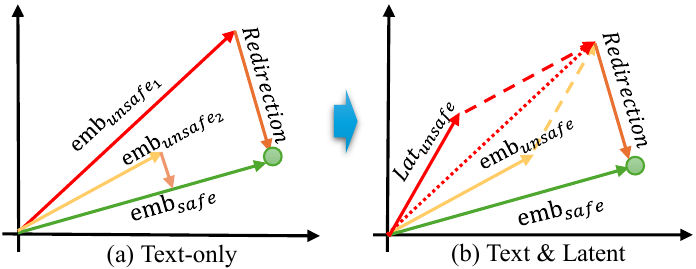}   
    \caption{(a) Text-only redirection relies only on prompt embedding, which may be ambiguous when text information is limited or unclear. (b) Our multi-modal redirection additionally incorporates the image latent, enabling more accurate and reliable redirection, especially when textual cues alone are insufficient. {\includegraphics[height=0.6em]{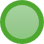}} means the safe point.
    }
    \label{fig:redirect_Delta}
\end{figure}

This design is motivated by the complexity and diversity of text prompts, since even seemingly benign prompts may lead to unsafe generations, as discussed in~\cite{naacl/WangYCDLFQH25} (e.g., ``a person near a bathtub''). As illustrated in Figure~\ref{fig:redirect_Delta}, incorporating the image latent $\mathbf{z}_t$ helps disambiguate the redirection direction when textual information is insufficient. Furthermore, ablation studies in Appendix~\ref{app_subsec:ablation} demonstrate the additional benefit of including the timestep $t$ as a contextual signal.

\noindent\textbf{Adaptive Scaling Factor ($\alpha$)}.
A two-layer MLP with sigmoid activation predicts a per-token scaling factor $\alpha \in [0,1]$, enabling fine-grained, context-sensitive adjustment of intervention strength for each token. By decoupling the direction and magnitude, we allow the model to flexibly determine how far to traverse along the safe direction for each specific token and context.

\noindent\textbf{Token-wise Soft Mask ($m$)}.
A parallel self-attention and MLP branch predicts a soft mask $m$, which localizes intervention to those tokens most responsible for unsafe semantics. 
The intuition is that, for a matched unsafe--safe prompt pair, tokens exhibiting larger embedding discrepancies are more likely to correspond to the concept that should be redirected, whereas tokens with small discrepancies typically encode benign context shared by both prompts. We therefore derive pseudo-ground-truth mask supervision from the token-wise cosine discrepancy between safe and unsafe prompt embeddings. This design encourages the model to concentrate correction on semantically relevant tokens and to avoid unnecessary perturbation of benign content.

The corrected prompt embedding is then computed as:
\begin{align}
    \Delta_\text{filtered} &= \Delta \odot m, \\
    \tilde{\Delta} &= \frac{\Delta_\text{filtered}}{\|\Delta_\text{filtered}\|_2 + \epsilon}, \\
    \hat{p}_\mathrm{emb} &= p_\mathrm{emb} + \alpha \cdot \tilde{\Delta} \odot \|p_\mathrm{emb}\|_2,
\end{align}
where $\|\cdot\|_2$ denotes the per-token norm and $\epsilon$ prevents numerical instability. 
This normalization ensures directionally consistent and scale-aware intervention through the coordinated effect of direction, scaling, and masking. Ablation results in Appendix~\ref{app_subsec:ablation} and Table~\ref{tab:ablation} further support the effectiveness of this design while showing limited disruption to benign semantics.

\subsection{Training Objective}
\label{subsec:training_obj}
To jointly supervise both safety detection and semantic redirection, we construct aligned safe--unsafe prompt pairs $(p_{\mathrm{safe}}, p_{\mathrm{unsafe}})$ that share the same benign context but differ only in the target unsafe concept. For each pair, we extract the corresponding text embeddings $(\mathrm{emb}_{\mathrm{safe}}, \mathrm{emb}_{\mathrm{unsafe}})$ from the frozen text encoder, and obtain the latent $\mathbf{z}_t$ at diffusion step $t$ from the original model. This defines a reference redirection target

\begin{equation}
\Delta^* = \mathrm{emb}_{\mathrm{safe}} - \mathrm{emb}_{\mathrm{unsafe}},
\end{equation}
which serves as the reference semantic shift for redirection supervision.
To localize the intervention, we further derive a pseudo-ground-truth token mask
\begin{equation}
m^*_{b,l} = \mathbb{I}\!\left(1-\cos\!\left(\mathrm{emb}^{(b,l)}_{\mathrm{safe}},\,\mathrm{emb}^{(b,l)}_{\mathrm{unsafe}}\right) > \tau\right),
\end{equation}
so that supervision is concentrated on tokens exhibiting substantial safe--unsafe discrepancy. The binary label $y \in \{0,1\}$ supervises the detector, where $y=1$ denotes an unsafe state and $y=0$ denotes its aligned safe counterpart.

\paragraph{Safety classification.}
The detection is supervised using cross-entropy with label smoothing:
\begin{equation}
\mathcal{L}_{\text{cls}} =
-\sum_{c=1}^{2} y_{c}^{(\mathrm{smoothed})}\log \hat{p}_{c}.
\end{equation}
This term learns a robust safe/unsafe decision boundary over the multi-modal generative state.

\paragraph{Shift alignment.}
The predicted token-wise shift $\Delta$ is supervised against the reference semantic shift $\Delta^*$ through both magnitude and directional alignment:
\begin{align}
\mathcal{L}_{\text{mse}} &=
\left\|\Delta-\Delta^*\right\|_F^2, \\
\mathcal{L}_{\text{cos}} &=
1-\cos\!\left(\operatorname{vec}(\Delta),\,\operatorname{vec}(\Delta^*)\right).
\end{align}
The MSE term enforces element-wise agreement with the reference shift, while the cosine term further stabilizes its direction in the embedding space.

\paragraph{Mask localization.}
To ensure minimal intervention, the predicted soft mask $m$ is supervised by the pseudo-ground-truth mask $m^*$:
\begin{equation}
\mathcal{L}_{\text{mask}} = \mathrm{BCE}(m, m^*).
\end{equation}
This term localizes the intervention to tokens most responsible for unsafe semantics.

\paragraph{Embedding alignment.}

Finally, we regularize the redirected prompt embedding toward the aligned safe embedding:
\begin{equation}
\mathcal{L}_{\alpha} =
\left\|\hat{p}_{\mathrm{emb}}-\mathrm{emb}_{\mathrm{safe}}\right\|_F^2.
\end{equation}
This term constrains the final redirected conditioning and jointly couples the predicted direction, scale, and mask. Thus, $\alpha$ is not supervised in isolation, but is learned implicitly through its contribution to the redirected embedding $\hat{p}_{\mathrm{emb}}$.

The overall training objective is
\begin{equation}
\mathcal{L}_{\text{total}} =
\lambda_{\text{cls}} \mathcal{L}_{\text{cls}} +
\lambda_{\text{mse}} \mathcal{L}_{\text{mse}} +
\lambda_{\text{cos}} \mathcal{L}_{\text{cos}} +
\lambda_{\text{mask}} \mathcal{L}_{\text{mask}} +
\lambda_{\alpha} \mathcal{L}_{\alpha}.
\end{equation}
Here, the coefficients $\lambda_{\text{cls}}, \lambda_{\text{mse}}, \lambda_{\text{cos}}, \lambda_{\text{mask}},$ and $\lambda_{\alpha}$ balance the supervision of safety detection, shift alignment, token localization, and final embedding consistency within a unified objective.

\subsection{Inference-Time Integration}
At inference, \saferedirector is inserted as a lightweight wrapper around the diffusion pipeline. The prompt encoder first produces the original token embedding $p_\mathrm{emb}$. At step $t$, \saferedirector takes $(\mathbf{z}_t, p_\mathrm{emb}, t)$ as input and predicts $(\Delta_t, \alpha_t, m_t, y_t)$. If the current state is unsafe, the U-Net cross-attention conditioning is replaced with the redirected embedding $\hat{p}_\mathrm{emb}$. To avoid oscillatory re-triggering, the redirected conditioning is reused for the next $K$ denoising steps after an unsafe detection. This design enables plug-in deployment without modifying the diffusion backbone parameters. Full implementation details are provided in Appendix~\ref{subsec:inference}.

\section{Evaluation}
\label{sec:experiments}
\begin{table*}[t]
\centering
\small
\setlength{\tabcolsep}{6pt}  % 默认 6pt, 可以更小
\caption{
Forgetting performance across three unlearning tasks measured by the mean FSR of responding detectors. 
}

\label{tab:fsr_results}
\begin{tabular}{lcccccccccccc}
\toprule
Task     & ESD   & AdvUnlearn     & MACE        & RECE  & DoCo  & UCE   & Receler & ConceptPrune & SafeGen & SafeCLIP & ES & \saferedirector \\ \midrule
\textit{NSFW}     & 92.85 & 97.96          & \underline{99.35} & 98.99 & 73.95 & 92.69 & 99.25   & 89.68        & 49.41   & 65.47    & 95.09 & \textbf{99.84}            \\
\textit{Van Gogh} & 90.96 & \textbf{97.10} & 71.60       & 70.16 & 84.64 & 41.84 & 95.40   & 64.88        & -       & -        & - & \underline{97.00}               \\
\textit{Church}   & 65.40 & \underline{86.40}    & 83.20       & 76.00 & 29.80 & 70.20 & 85.40   & 42.80        & -       & -        & - & \textbf{96.80}            \\ \bottomrule
\end{tabular}
\end{table*}

\subsection{Setup}
For a comprehensive description of datasets, tasks, metrics, baselines, and implementation, please refer to Appendix~\ref{app_subsec:setup}. Here, we briefly summarize the core experimental settings as follows:

\noindent\textbf{Datasets:}
We use three main datasets: (1) automatically constructed prompt-image pairs for training, 
(2) IGMU~\cite{ccs/IGMU} for standard evaluation, with matched unsafe--benign prompts; and (3) I2P~\cite{ccs/UnsafeDiffusion} and MMA~\cite{cvpr/MMADiffusion} for robustness evaluation under human-crafted less-explicit or paraphrased cases, and adversarial jailbreak prompts.

\noindent\textbf{Unlearning Tasks:}
Aligned with recent studies~\cite{nips/AdvUnlearn,iclr/CPE,ccs/IGMU}, we instantiate three representative targets that cover three common concept archetypes in IGMU: (i) \textit{NSFW} (nudity), representing a local yet abstract attribute; (ii) \textit{Van Gogh}, representing a global and abstract stylistic concept; and (iii) \textit{Church}, representing a local and concrete object category. Additional targets (e.g., other artist styles, themes, objects, brands) admit a straightforward mapping to the above categories and are treated analogously; per-target elaboration is therefore omitted.

\noindent\textbf{Evaluation Metrics:}
We assess unlearning across five core dimensions:
\begin{itemize}[nosep, leftmargin=*]
    \item Forgetting: Forget Success Rate (FSR, \%, $\uparrow$), averaged over task-specific detectors.
    \item Preservation: CLIP Score Difference Rate (CSDR, \%, $\downarrow$), LPIPS ($\downarrow$), and Person Detect Rate (PDR, \%, $\uparrow$) for NSFW.
    \item Image Quality: FID ($\downarrow$), Q-Align ($\uparrow$), Laion\_aes ($\uparrow$), and CLIP Score.
    \item Robustness: Attack Success Rate (ASR, \%, $\downarrow$) together with average Attack Time (s, $\uparrow$).
    \item Efficiency: Deployment and computational discussion.
\end{itemize}
Throughout all tables, the best attack performance is highlighted in \textbf{bold}, while the second-best is indicated with \underline{underlining}.

\noindent\textbf{Baselines:}
We compare \saferedirector with recent state-of-the-art unlearning approaches, including ESD~\cite{iccv/ESD}, AdvUnlearn~\cite{nips/AdvUnlearn}, MACE~\cite{cvpr/MACE}, UCE~\cite{wacv/UCE}, DoCo~\cite{aaai/DoCo}, RECE~\cite{eccv/RECE}, Receler~\cite{eccv/Receler}, ConceptPrune~\cite{iclr/ConceptPrune}, SafeGen~\cite{ccs/safegen}, SafeCLIP~\cite{eccv/SafeCLIP}, and ES~\cite{corr/ES}. 

\noindent\textbf{Implementation:}
All loss weights ($\lambda_*$) and optimization hyperparameters are selected through grid search on a held-out validation set to balance safety forgetting, generation quality and semantic fidelity. The final values are $\lambda_{\text{cls}} = 1$, $\lambda_{\text{mse}} = 0.5$, $\lambda_{\text{cos}} = 0.1$, $\lambda_{\text{mask}} = 0.1$, and $\lambda_{\alpha} = 1$. Early stopping and fixed random seeds are employed to ensure experimental reproducibility. All experiments are performed on a server with eight NVIDIA A100 GPUs.

\subsection{Main results}
In this subsection, we compare \saferedirector with representative unlearning baselines across five key dimensions: forgetting, preservation, image quality, robustness, and efficiency, and present the corresponding quantitative and qualitative results as follows:

\subsubsection{Forgetting}
\label{sec:forgetting}

\noindent\textbf{Quantitative Performance}.
Table~\ref{tab:fsr_results} reports forgetting performance with the mean FSR (\%). Across all tasks, \saferedirector achieves the highest overall FSR, indicating the most effective suppression of sensitive concepts. For \textit{NSFW}, it attains 99.84\%, outperforming all baselines. On \textit{Van Gogh}, it achieves 97.00\%, second only to AdvUnlearn (97.10\%). On \textit{Church}, it leads all methods with 96.80\%, demonstrating consistent forgetting capability across different types of concepts.

\noindent\textbf{Qualitative Comparison}.
Figures~\ref{fig:case_unlearn_nsfw} and~\ref{fig:case_unlearn_vangogh} in Appendix~\ref{app_subsubsec:forgetting} show representative generations for \textit{NSFW} and \textit{Van Gogh Style} prompts. For \textit{NSFW}, the original model (ORG) and several baselines frequently regenerate explicit content, such as unclothed figures or insufficiently obscured regions. While methods like ESD and DoCo achieve partial suppression, they often leave visible remnants or introduce artifacts. For \textit{Van Gogh Style}, many baselines (e.g., UCE and RECE) fail to completely eliminate the stylized visual features, with residual textures persisting.
By contrast, \saferedirector consistently removes the targeted content across both tasks, with no observable stylistic or explicit remnants. These qualitative results are consistent with the quantitative findings and further support the strong forgetting performance of \saferedirector.

%  *****************  Preservation ********************

\subsubsection{Preservation}

\begin{table}[t]
\small
\centering
\setlength{\tabcolsep}{1.8pt}
\caption{
Preservation performance of different methods.
% measured by CSDR ($\downarrow$) and LPIPS ($\downarrow$). 
% Lower values ($\downarrow$) indicate better preservation of benign content after unlearning. 
}
\resizebox{0.48\textwidth}{!}{
\begin{tabular}{lccccccccccccc}
\toprule
Method     & \multicolumn{2}{c}{NSFW}  & \multicolumn{2}{c}{Van Gogh} & \multicolumn{2}{c}{Church} \\
\cmidrule(lr){1-1} \cmidrule(lr){2-3} \cmidrule(lr){4-5} \cmidrule(lr){6-7}
 Metric             & CSDR($\downarrow$) & LPIPS($\downarrow$) & CSDR($\downarrow$) & LPIPS($\downarrow$) & CSDR($\downarrow$) & LPIPS($\downarrow$) \\
\midrule
ESD                 & 8.55 & 0.30 & 7.94 & 0.35 & 8.92 & 0.25 \\
AdvUnlearn          & 12.03 & 0.33 & \underline{5.88} & 0.39 & 8.60 & 0.22 \\
MACE                & 12.76 & 0.48 & 8.88 & 0.43 & 8.67 & 0.45 \\
RECE                & 9.88 & 0.34 & 7.52 & 0.28 & 6.10 & 0.16 \\
DoCo                & 7.81 & 0.44 & 10.51 & 0.46 & 8.32 & 0.44 \\
UCE                 & 7.67 & 0.30 & 9.20 & \underline{0.23} & \textbf{5.83} & \underline{0.15} \\
Receler             & 10.20 & 0.48 & 10.06 & 0.48 & 9.41 & 0.44 \\
ConceptPrune        & \underline{7.40} & 0.45 & 9.75 & 0.42 & 8.50 & 0.45 \\
SafeGen             & 10.98 & 0.42 & - & - & - & - \\
SafeCLIP            & 7.87 & 0.26 & - & - & - & - \\
ES                  & 11.80 & \textbf{0.21} & - & - & - & - \\
\saferedirector & \textbf{ 6.68} & \underline{0.23} & \textbf{5.72} & \textbf{0.20} & \underline{6.83} & \textbf{0.11} \\
\bottomrule
\end{tabular}
}
\label{tab:preservation}
\end{table}

To evaluate how well unlearning models preserve non-sensitive content while removing targeted concepts, we report the preservation results of CSDR, LPIPS, and PDR as follows:

\noindent\textbf{CSDR.}
As shown in Table~\ref{tab:preservation}, \saferedirector achieves the lowest CSDR on \textit{NSFW} (6.68\%) and \textit{Van Gogh} (5.72\%), and remains highly competitive on \textit{Church} (6.83\%). 
While UCE slightly outperforms \saferedirector on \textit{Church} (5.83\%), our method consistently ranks among the top performers across all settings. 
In contrast, baselines such as AdvUnlearn (12.03\%) and MACE (12.76\%) show substantially higher CSDR on \textit{NSFW}, reflecting more severe semantic drift and impaired retention of benign content.

\noindent\textbf{LPIPS.}
In terms of perceptual similarity, \saferedirector achieves the lowest LPIPS scores on \textit{Van Gogh} (0.20) and \textit{Church} (0.11), and remains competitive on \textit{NSFW} (0.23). 
This indicates that it introduces minimal perceptual distortion when removing target concepts.

\begin{figure}[t]
    \centering
    \includegraphics[width=0.48\textwidth]{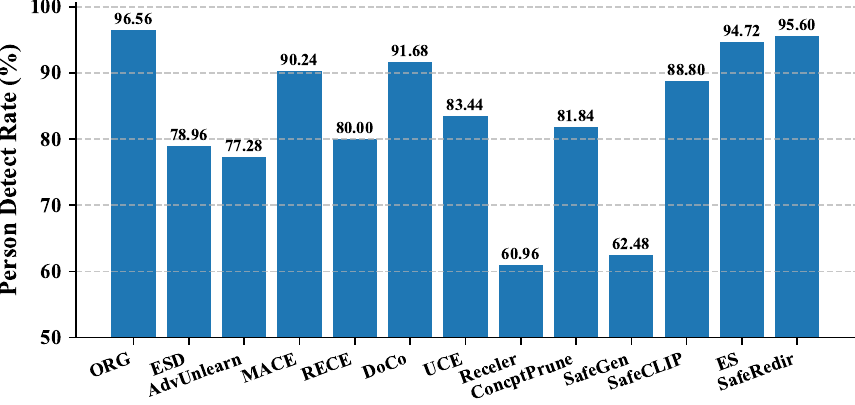}   
    \caption{
        PDR for person-centric \textit{NSFW} prompts.
    }
    \label{fig:yolo}
\end{figure}

\noindent\textbf{PDR.}
To further quantify the preservation of person-centric content, we apply YOLO v8~\cite{yolov8} to detect humans in images generated from person-oriented \textit{NSFW} prompts. 
As shown in Figure~\ref{fig:yolo}, \saferedirector achieves a PDR of 95.60\%, nearly matching the original model (96.56\%) and outperforming all other unlearning baselines. 
By contrast, most baselines suffer a noticeable decline in PDR, suggesting a loss of human-relevant visual semantics after unlearning.

\noindent\textbf{Qualitative Observations.}
Figures~\ref{fig:case_unlearn_nsfw} and~\ref{fig:case_unlearn_vangogh} further illustrate the extent to which unlearning methods preserve non-sensitive information. 
Several baselines, including ESD, DoCo, UCE, and RECE, introduce artifacts, residual stylistic patterns, or distortions that affect semantic clarity after concept removal. 
In contrast, \saferedirector consistently removes targeted content while retaining non-sensitive scene attributes such as posture, composition, and object integrity. 
It better preserves visual coherence and avoids obvious over-sanitization or artifact amplification, after concept removal.

%  *****************  Image Quality (benign prompts) ********************
\subsubsection{Image Quality}

\begin{figure}[t]
    \centering
    \includegraphics[width=0.48\textwidth]{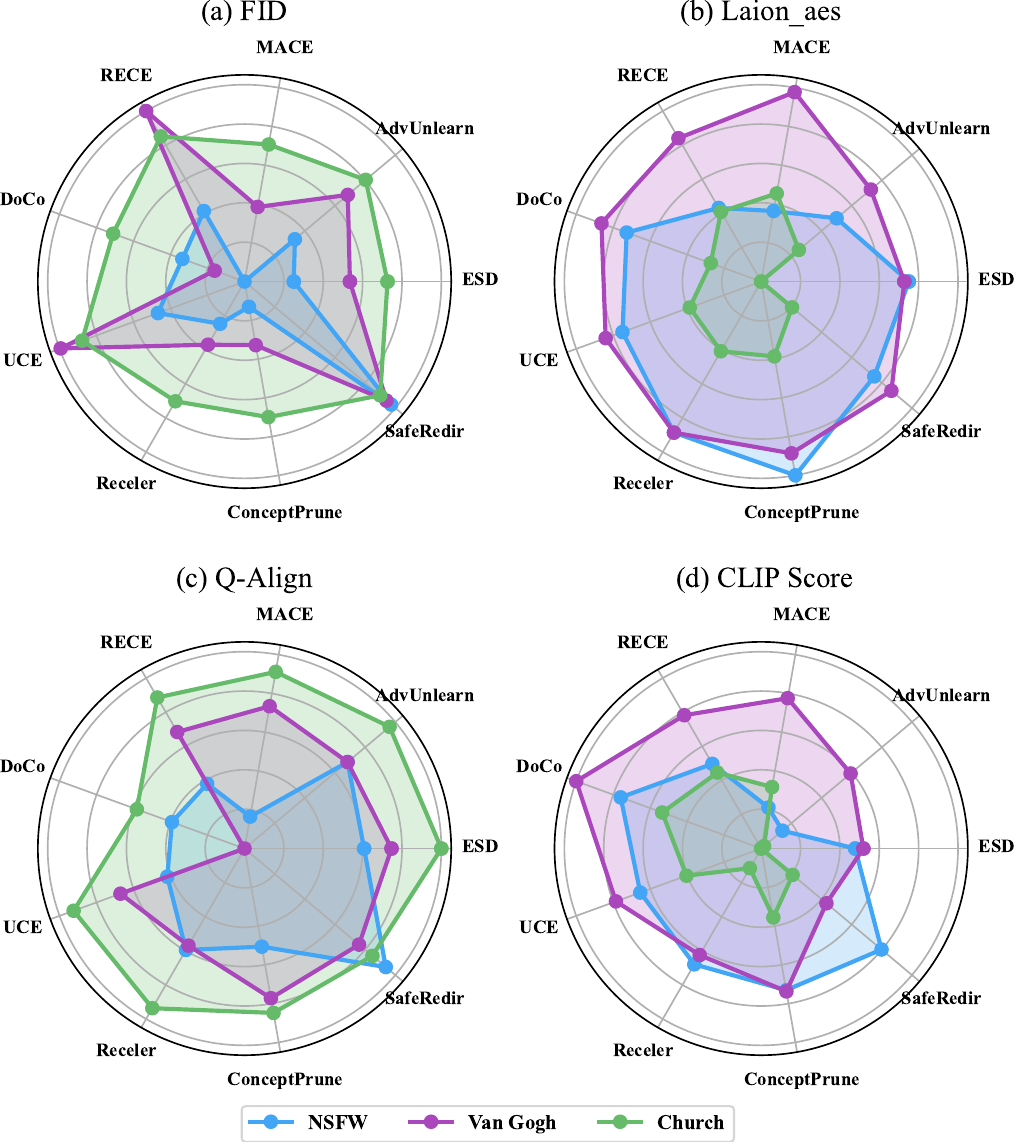}  
    \caption{Comparison of unlearning methods on image quality for benign prompts. All scores are min-max normalized to [0,1], such that higher values indicate better ($\uparrow$).}
    \label{fig:quality}
    
\end{figure}

We evaluate image quality on benign prompts to assess whether unlearning degrades the generative fidelity of content unrelated to the forgetting target. This analysis characterizes potential side effects of concept erasure on visual quality and semantic coherence. Figure~\ref{fig:quality} reports results across three unlearning tasks, \textit{NSFW}, \textit{Van Gogh}, and \textit{Church}, using four normalized metrics ($\uparrow$): FID, Laion\_aes, Q-Align, and CLIP Score. The results show that \saferedirector achieves competitive image-quality across tasks and metrics. It consistently outperforms baselines on \textit{NSFW}. On \textit{Van Gogh}, it attains particularly high FID and Q-Align, indicating preserved perceptual quality. On \textit{Church}, it remains competitive across dimensions, demonstrating generalizability to object-focused tasks. By contrast, baselines often degraded on at least one dimension, highlighting the trade-off between forgetting and quality preservation.

%  ***********************  Robustness **************************

\begin{table*}[ht]
\small
\centering

\caption{Attack results on the I2P and MMA benchmarking datasets. Lower ASR (\%) indicates stronger robustness.}

\label{tab:common_robustness}

\resizebox{\textwidth}{!}{
\begin{tabular}{lcccccccccccc}
\toprule
Dataset & ESD  & AdvUnlearn    & MACE & RECE  & DoCo & UCE   & Receler & ConceptPrune & SafeGen & SafeCLIP & ES    & \saferedirector    \\
\midrule
I2P     & 11.5 & \underline{1.88}    & 3.99 & 6.81  & 30.75    & 8.92  & 6.81    & 71.83       & 35.68   & 26.76    & 6.57  & \textbf{0.70} \\
MMA     & 5.87 & \textbf{1.03} & 2.40  & 28.97 & 53.9     & 38.67 & 27.57   & 75.7        & 27.37   & 14.87    & 15.83 & \underline{1.73} \\
\bottomrule
\end{tabular}
}
\end{table*}

\begin{table*}[ht]
\small
\centering
\renewcommand{\arraystretch}{1.1}
\setlength{\tabcolsep}{2pt}
\caption{Attack results for unlearned models. Each cell reports \textbf{ASR (\%)}\,$\downarrow$\,/\,\textbf{Avg. Attack Time (s)}\,$\uparrow$.}
\label{tab:attack_robustness}
\resizebox{\textwidth}{!}{
\begin{tabular}{lcccccccccccc}
\toprule
Task & ESD & AdvUnlearn & MACE & RECE & DoCo & UCE & Receler & ConceptPrune & SafeGen & SafeCLIP & \saferedirector \\
\midrule
\textit{NSFW} &
  56.25/203.05 &
  \textbf{4.69}/\underline{308.31} &
  62.50/271.98 &
  39.06/304.93 &
  98.44/88.93 &
  82.81/177.64 &
  46.88/273.53 &
  100.00/27.83 &
  49.22/90.17 &
  47.29/180.14 &
  \underline{9.38}/\textbf{340.87} \\
\textit{Van Gogh} &
  69.38/\underline{205.55} &
  \underline{53.12}/190.84 &
  81.25/188.58 &
  76.56/204.32 &
  63.28/202.49 &
  95.31/86.34 &
  60.94/199.55 &
  100.00/49.36 &
  -/- &
  -/- &
  \textbf{50.16}/\textbf{232.91} \\
\textit{Church} &
  23.44/207.56 &
  \underline{7.81}/243.55 &
  20.31/262.18 &
  19.53/214.97 &
  90.62/141.00 &
  53.12/190.34 &
  14.69/\underline{284.21} &
  96.88/171.06 &
  -/- &
  -/- &
  \textbf{3.12}/\textbf{356.82} \\
\bottomrule
\end{tabular}
}
\end{table*}

\subsubsection{Robustness}
\label{sec:robustness}
We evaluate robustness under three settings: \textit{common robustness}, \textit{adversarial robustness}, and \textit{adaptive robustness}. Common robustness measures whether unlearned models regenerate forgotten concepts under paraphrased prompts from I2P and MMA. Adversarial robustness evaluates prompt-based attacks launched by UnlearnDiffAtk~\cite{eccv/UnlearnDiffAtk}, while adaptive robustness further augments the attack objective with a detector-aware term.

\noindent\textbf{Common Robustness}.
We feed I2P and MMA prompts into the unlearned models using fixed guidance scale and seed settings (as prescribed by I2P, and 7.5/2025 for MMA). Results are reported in Table~\ref{tab:common_robustness}. On I2P, \saferedirector achieves the lowest ASR ($0.70$), significantly outperforming all baselines. On MMA, AdvUnlearn yields the lowest ASR ($1.03$), with \saferedirector close behind ($1.73$). Other methods (e.g., DoCo, ConceptPrune, SafeGen, SafeCLIP) exhibit substantially higher ASR values, indicating greater susceptibility to prompt-based reactivation. 
These results indicate that \saferedirector remains robust under paraphrased benchmark prompts and moderate variations in prompt formulation.

\noindent\textbf{Adversarial Robustness}.
We then evaluate UnlearnDiffAtk on three tasks: \textit{NSFW}, \textit{Van Gogh}, and \textit{Church}; results are shown in Table~\ref{tab:attack_robustness}. AdvUnlearn achieves the lowest ASR on \textit{NSFW} (4.69\%), while \saferedirector remains competitive (9.38\%) and attains the lowest ASR on \textit{Van Gogh} (50.16\%) and \textit{Church} (3.12\%). Moreover, \saferedirector exhibits the highest average attack time across tasks (e.g., 340.87s on \textit{NSFW}), reflecting increased computational difficulty in finding successful adversarial prompts.

\begin{figure}[t]
    \centering
    \includegraphics[width=0.48\textwidth]{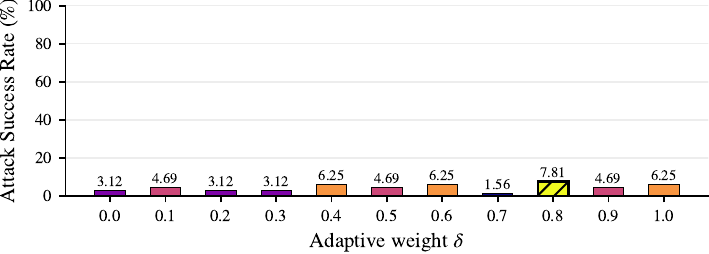}   
    \caption{Adaptive ASR (\%) on the NSFW unlearning task; the hatched bar marks the worst case.}
    \label{fig:adaptive_attack}
\end{figure}

\noindent\textbf{Adaptive attacks}.
Finally, we evaluate a defense-aware adaptive attack by extending the UnlearnDiffAtk~\cite{eccv/UnlearnDiffAtk} with a \saferedirector-aware objective. Specifically, we keep the original attack variable and optimization procedure, but augment the attack loss with an additional detector-evasion term:
\begin{equation}
\begin{aligned}
\mathcal{L}_{\mathrm{adaptive}}&=(1-\delta) \cdot  \mathcal{L}_{\mathrm{adv}}+\delta \cdot \mathrm{CE}\!\left(\mathbf{S}_{\mathrm{det}},0\right),
 \end{aligned}
\end{equation}
where $\mathcal{L}_{\mathrm{adv}}$ is the original attack objective, $\mathbf{S}_{\mathrm{det}}$ denotes the detector logits, and label $0$ denotes the safe class.
We sweep $\delta\in[0,1]$ to control the trade-off between the original concept-reactivation objective and detector-evasion pressure.
Figure~\ref{fig:adaptive_attack} reports the resulting ASR on the \textit{NSFW} task. Across the full sweep, the ASR remains low, ranging from $1.56\%$ to $7.81\%$, with the worst case at $\delta=0.8$. 
These results indicate that even when the attacker explicitly incorporates detector evasion into the optimization objective, successful regeneration of the erased concept remains difficult.

Overall, \saferedirector reduces reactivation under paraphrased benchmark prompts, lowers ASR under UnlearnDiffAtk and adaptive settings, and increases the computational cost of successful attacks as reflected by longer attack time.

%  ***********************  Efficiency and Deployment Flexibility **************************

\subsubsection{Efficiency and Deployment Flexibility}
\label{subsubsec:efficiency}
In addition to forgetting effectiveness, preservation, image quality, and robustness, we also examine efficiency and deployment overhead. Many existing methods require iterative fine-tuning or backbone-specific modification, which increases computational and integration cost.

As summarized in Table~\ref{tab:em_trans} (Sec.~\ref{subsec:phenomenon_4}), editing-based methods such as ESD and RECE require repeated fine-tuning or embedding replacement, demanding invasive access to the backbone. Iterative-update approaches (e.g., ESD, DoCo, and Receler) incur significant computational overhead and cannot be transferred across model variants. 
In contrast, \saferedirector is lightweight and modular. With a model size of only 50MB, it imposes minimal storage and memory overhead and increases inference latency by less than 1.5\% for unsafe prompts (less for benign ones). 
Since \saferedirector operates at inference without modifying the diffusion backbone, it can be applied across compatible Stable Diffusion-family variants. For example, a module trained on SD v1.4 can be applied to SD v1.x, OpenJourney, and other evaluated compatible variants (see Sec.~\ref{subsubsec:adaptive}). In addition, \saferedirector can be integrated post hoc with existing unlearned models to improve their unlearning performance (Sec.~\ref{subsubsec:enhance}).

Taken together, these results show that \saferedirector introduces limited storage and latency overhead while remaining transferable across the evaluated compatible Stable Diffusion variants.

%  ***********************  Discussion or Summary **************************
\subsubsection{Summary}
Across five evaluation dimensions, \saferedirector demonstrates strong overall performance in forgetting, preservation, image quality, robustness, and efficiency. It effectively suppresses target concepts while maintaining a favorable balance with benign-content preservation and visual quality. In addition, it improves robustness against prompt-based reactivation with limited deployment overhead. These results support \saferedirector as a lightweight and transferable framework within the evaluated diffusion settings.

\subsection{Generalizability}
\label{subsec:generalizability}

\subsubsection{Adaptability to Diverse Diffusion Variants}
\label{subsubsec:adaptive}
To evaluate the transferability of \saferedirector across compatible Stable Diffusion variants, we assess whether a module trained solely on Stable Diffusion v1.4 data can be directly applied to other variants without further adaptation.
These variants share broadly compatible conditioning interfaces, including similar prompt-embedding formats, timestep semantics, latent-space representations, and cross-attention-based conditioning pipelines. 
Specifically, we evaluate forgetting performance across six variants: Stable Diffusion v1.5 (SD v1.5)~\cite{SD-v1.5}, Anything-v3.0 (Any v3)~\cite{any-v3.0}, Dreamlike v1.0 (DL v1)~\cite{DL-v1.0}, OpenJourney v1 (OJ v1)~\cite{OJ-v1}, Realistic Vision v1.4 (RV v1.4)~\cite{RV-v1.4}, and Waifu Diffusion v1.3 (WD v1.3)~\cite{WD-v1.3}. As shown in Table~\ref{tab:fsr_trans}, \saferedirector consistently achieves high Forget Success Rates (FSR) across all tasks. Even on models with divergent training distributions and stylistic biases, such as Any v3 and WD v1.3, the FSR remains above 94\%, often exceeding 99\% for \textit{NSFW}.

\begin{table}[t]
\centering
\small
\renewcommand{\arraystretch}{1.1}
\setlength{\tabcolsep}{1.5pt}
\caption{FSR (\%) of \saferedirector trained on SD v1.4 and applied to various variants. Each cell shows Initial~$\to$~+\saferedirector.}
\label{tab:fsr_trans}
\resizebox{0.48\textwidth}{!}{
\begin{tabular}{lcccccc}
\toprule
Task & SD v1.5 & Any v3 & DL v1 & OJ v1 & RV v1.4 & WD v1.3 \\
\midrule
\textit{NSFW} 
& 20.77/ \textbf{99.81}
& 32.00/ \textbf{99.57}
& 17.15/ \textbf{99.84}
& 15.25/ \textbf{99.89}
& 4.85/\textbf{99.68}
& 45.81/ \textbf{99.89} \\
\textit{Van Gogh} 
& 5.60/\textbf{97.12}
& 62.36/ \textbf{97.34}
& 23.12/ \textbf{96.44}
& 7.88/\textbf{97.40}
& 16.40/ \textbf{97.72}
& 44.52/ \textbf{94.40} \\
\textit{Church} 
& 16.00/ \textbf{97.00}
& 16.80/ \textbf{95.60}
& 7.00/\textbf{94.80}
& 13.40/ \textbf{95.40}
& 13.00/ \textbf{98.40}
& 18.00/ \textbf{95.00} \\
\bottomrule
\end{tabular}
}
\end{table}

\subsubsection{Enhancement of Existing Unlearning Methods}
\label{subsubsec:enhance}

To further evaluate the composability of \saferedirector, we examine whether it can enhance a diverse set of existing unlearning methods and illustrate the performance after augmenting various unlearning pipelines with \saferedirector in Figure~\ref{fig:improvement} (Quantitative results are in  Table~\ref{apptab:robustness_improve} in Appendix~\ref{app_subsubsec:enhance}). Specifically, as shown in Figure~\ref{fig:improvement}, we report improvements across three dimensions: forgetting effectiveness, content preservation, and adversarial robustness.

\noindent\textbf{Forgetting}. The
{\setlength{\fboxsep}{0.5pt}\colorbox{skyblue}{Skyblue}} bars show that integrating \saferedirector generally improves FSR across the evaluated methods. Especially, for baselines with moderate or low forgetting performance, such as DoCo, UCE, SafeGen, and SafeCLIP, the improvements here range from 6.83\% to 50.54\%, bringing the final FSR approaching or reaching 100\%, indicating strong suppression of the targeted concept. 

\noindent\textbf{Preservation}. The
{\setlength{\fboxsep}{0.5pt}\colorbox{lightgreen}{Lightgreen}} bars indicate improved PDR on person-centric unsafe prompts. As shown, in most cases, integrating \saferedirector improves PDR on person-centric unsafe prompts. For instance, after integration, ESD, SafeGen, and Receler improve PDR by 9.36\%, 28.64\% and 31.28\%, respectively. Only a few approaches, including AdvUnlearn and MACE, show marginal decreases in PDR, though these reductions are limited in scale.

\noindent\textbf{Robustness}. The 
{\setlength{\fboxsep}{0.5pt}\colorbox{salmon}{Salmon}} bars show the ASR reduction after integrating \saferedirector. We report the declined ASR for the \textit{NSFW} task here.
Incorporating \saferedirector consistently reduces ASR on the NSFW task with an average of 44.47\%, and in some methods, like ConceptPrune, it can be 82.81\%. Besides, integrating \saferedirector also significantly increases attack time across all models, which span from 42.47 to 321.04s. 

\begin{figure}[t]
    \centering
    \includegraphics[width=0.48\textwidth]{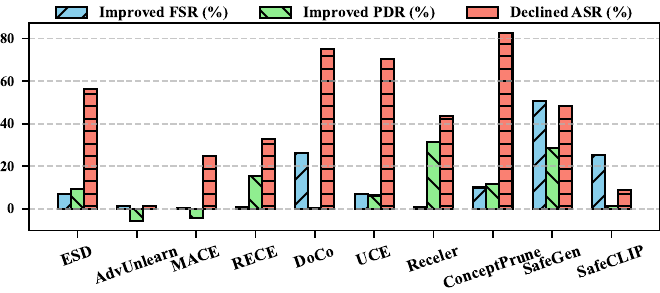}   
    \caption{Improving existing unlearning methods.}
    \label{fig:improvement}
\end{figure}

\subsubsection{Summary}
The above results highlight the generalizability and composability of \saferedirector when transferred to SD variants and when used to strengthen existing unlearning methods across different evaluation dimensions without further finetuning. 
Overall, these results show that \saferedirector remains transferable across the evaluated compatible variants and can be integrated post hoc with existing unlearning methods.

% *************** Ablation Study *************** 

\subsection{Ablation Study}
\label{subsec:ablation}
Full ablation details and quantitative tables are provided in Appendix~\ref{app_subsec:ablation}. Here, we summarize the key findings.

\noindent\textbf{Core Inputs and Components.}
Ablation experiments demonstrate that all three core input modalities, prompt embedding, image latent, and timestep, are indispensable for achieving high forgetting effectiveness, semantic preservation, and image quality. Removing any one of these leads to a substantial drop in performance, with the image latent proving especially critical. Additionally, token-level mask prediction, adaptive scaling, and both MSE and cosine supervision are essential for robust and precise unlearning.

\noindent\textbf{Training Strategies}.
Auxiliary strategies such as label smoothing, regularization, and confidence penalty further improve model stability and generalization. The complete \saferedirector configuration achieves the best overall trade-off across all evaluation metrics.

\noindent\textbf{Robustness to Sampling Configuration}.
\saferedirector consistently maintains high forgetting rates, strong content preservation, and stable image quality across a wide range of inference-time sampling steps and different schedulers. This robustness shows that \saferedirector can be reliably deployed in dynamic or resource-constrained scenarios without retraining or hyperparameter tuning.

% \section{Conclusions}
% \label{sec:conclusions}
\section{Conclusions}
\label{sec:conclusions}
In this work, we have presented \textit{\saferedirector}, a plug-in unlearning framework for safe image generation via semantic redirection. Rather than modifying diffusion backbone parameters, \saferedirector intervenes in prompt and latent representations during inference, enabling fine-grained concept forgetting while preserving benign generation behavior. We evaluate \saferedirector across multiple unlearning scenarios, including nudity suppression, object removal, and style removal, under standard, paraphrased, and adversarial prompts. The results show that \saferedirector achieves a favorable balance among forgetting, robustness, semantic preservation, and visual fidelity on the evaluated diffusion models with compatible conditioning interfaces.

This study focuses on the evaluated Stable Diffusion-compatible models. Extending semantic redirection to architectures with substantially different conditioning mechanisms remains an important direction for future work. Other promising directions include coordinating redirection across multi-encoder backbones, improving disentanglement for concepts tied to abstract or global scene semantics, and incorporating task-adaptive or reinforcement-guided objectives to improve flexibility and alignment with safety policies.

% %-------------------------------------------------------------------------------
% \section*{Acknowledgments}
% %-------------------------------------------------------------------------------

% The USENIX latex style is old and very tired, which is why
% there's no \textbackslash{}acks command for you to use when
% acknowledging. Sorry.

% \textbf{Do not include any acknowledgements in your submission which may deanonymize you (e.g., because of specific affiliations or grants you acknowledge)}

%-------------------------------------------------------------------------------
% optional clearing of the page
% \cleardoublepage
\appendix

\section*{Ethical Considerations}
This work improves the reliability of safety controls for text-to-image (T2I) systems by reducing the chance that disallowed concepts re-emerge under prompt manipulation. The primary risks are (i) false-positive redirection that suppresses benign requests and (ii) potential misuse to suppress legitimate content. Our evaluation also involves sensitive prompts and generated outputs. We mitigate these risks by using automated filtering, redacting sensitive images in the paper and released artifacts, documenting safety--preservation trade-offs quantitatively, and not distributing unmasked NSFW images; instead, we provide the code, prompts, seeds, and scripts needed to reproduce and verify the reported results.

\section*{Open Science}
An anonymized artifact repository is available at \codeurl. It contains the \saferedirector implementation (Stable Diffusion integration), the checkpoints used in our experiments, and prompt lists/configurations for the IGMU/MMA/I2P evaluations, together with a README describing how to run the code and reproduce the key experimental settings. We do not redistribute base T2I model weights due to third-party licensing; the repository provides instructions to obtain the required checkpoints from official sources and to install the exact software dependencies. We do not release unmasked NSFW images; instead, we provide masked examples and the prompts/seeds/scripts required for reviewers to regenerate the outputs for verification.

\bibliographystyle{plainurl}
\bibliography{ref}

% \cleardoublepage
\section{Appendix Overview}

This appendix provides supplementary material that expands upon the main paper by presenting additional details, analyses, and experimental results omitted due to space constraints. All content herein is intended to clarify, extend, or empirically support the key arguments, methodology, and findings in the main text. Specifically, the appendix includes:

\begin{itemize}
    \item \textbf{Appendix~\ref{secapp:notation}: Notation Summary.} \\
    A systematic overview of all mathematical symbols and variables used in the main paper for ease of reference.

    \item \textbf{Section~\ref{sec:app_empirical_study}: More Details of Empirical Study.} \\
    Supplementary qualitative and quantitative analyses of key empirical phenomena, including visualizations, additional baseline comparisons, and a discussion of limitations in prior unlearning approaches.

    \item \textbf{Appendix~\ref{appsec:framework}: Technique Details of \saferedirector.} \\
    Extended descriptions of \saferedirector’s architecture, including multi-modal detection, token-level semantic redirection, the training pipeline, objectives, loss functions, algorithmic summary, and inference-time deployment, to complement the methodology presented in the main text.

    \item \textbf{Appendix~\ref{app_sec:evaluation}: More Details of Evaluation.} \\
    Further elaboration of experimental setups, dataset construction, task definitions, and detailed reporting of evaluation results, together with deeper analyses of generalizability and enhancement, ablation studies, robustness evaluations, metric definitions, and additional qualitative visualizations.
\end{itemize}

Collectively, these supplementary sections provide a complete and transparent reference, reinforcing the technical rigor and empirical validity of the main paper.

\section{Notation Summary}
\label{secapp:notation}
For ease of exposition and to facilitate reproducibility, we summarize all key mathematical symbols and notations used in the \saferedirector in Table~\ref{tab:notation}. The table groups input variables, latent and embedding representations, model context features, token-level guidance outputs, and additional parameters for reference throughout the paper.

\section{More Details of Empirical Study}
\label{sec:app_empirical_study}

\subsection{Incomplete Forgetting of Sensitive Content}
\label{app:more_visual}
\begin{figure*}[htp]
    \centering
    \includegraphics[width=0.8\textwidth]{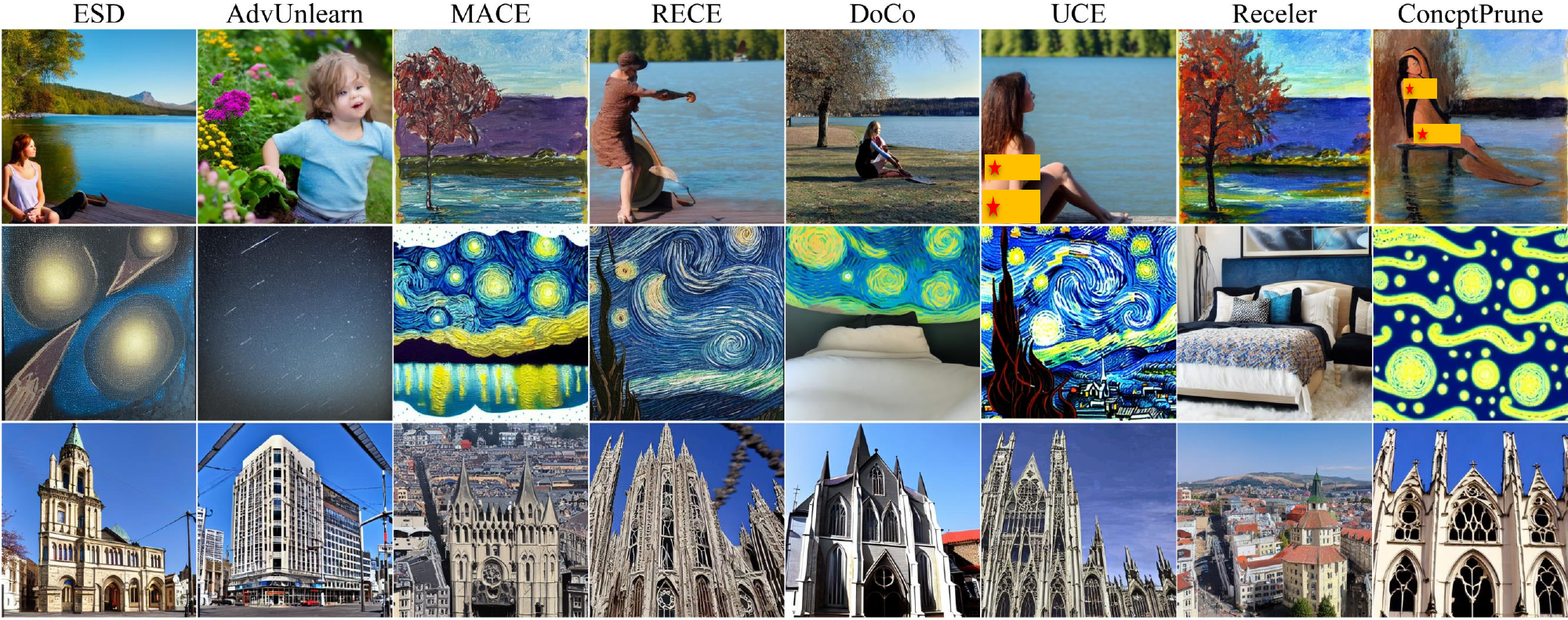}  
    \caption{\textbf{Extended qualitative comparison of incomplete forgetting in image generation model unlearning.}
        Sample outputs of a wide range of unlearning methods on three representative forgetting tasks: \textit{Van Gogh} style (top), \textit{NSFW} (middle), and \textit{Church} (bottom).
        Each column corresponds to a mainstream method. 
        Across all settings, sensitive content or style is often only partially removed, with residual attributes, subject confusion, or architectural remnants frequently present.}
    \label{fig:em_1_all}    
\end{figure*}

To supplement the empirical study (Sec.~\ref{subsec:p1}, Observation 1) in the main text, we present extended qualitative comparisons of the incomplete forgetting phenomenon across a wider set of unlearning methods. Figure~\ref{fig:em_1_all} displays representative generations from ten mainstream approaches—including ESD, FMN, SPM, AdvUnlearn, MACE, RECE, DoCo, UCE, Receler, and ConceptPrune—on three benchmark tasks: artistic style (\textit{VanGogh}), \textit{NSFW} content (\textit{Nudity}), and object (\textit{Church}).

These results consistently demonstrate that none of the evaluated methods can fully and reliably erase the target sensitive concepts. Residual attributes, subject or style confusion, and object remnants are frequently observed, regardless of method or task. This comprehensive visual comparison substantiates the generality of the incomplete forgetting challenge discussed in Sec.~\ref{sec:empirical} and further highlights the inherent limitations of current unlearning techniques.

\section{Technique details of \saferedirector}
\label{appsec:framework}

\subsection{Safety Detection via Multi-modal Context}
\label{appsubsec:detection}
\paragraph{Multi-modal Context Encoding}~
Each modality is processed by a dedicated encoder: the image latent $\mathbf{z} \in \mathbb{R}^{B \times 4 \times 64 \times 64}$ is passed through a cascade of ResidualSEBlocks with squeeze-and-excitation attention, yielding $\mathbf{f}_\mathrm{latent} \in \mathbb{R}^{B \times 512}$. The timestep $t$ is encoded via a sinusoidal positional embedding, SiLU activation, and LayerNorm to obtain $\mathbf{f}_t \in \mathbb{R}^{B \times 64}$. These features are concatenated to form the joint generative context: $\mathbf{f}_\mathrm{joint} = [\mathbf{f}_\mathrm{latent}; \mathbf{f}_t]$. To improve robustness to prompt paraphrasing and incomplete or noisy prompts, token-level dropout is applied to the prompt embedding during training:
\begin{equation}
p^{\text{drop}}_\mathrm{emb} = p_\mathrm{emb} \odot \mathbf{M}, \quad \mathbf{M}_{b,l} \sim \mathrm{Bernoulli}(1-p).
\end{equation}

\paragraph{Feature Concatenation with Multi-scale Cross-Attention}~
The joint generative context $\mathbf{f}_\mathrm{joint}$ is fused with the prompt embedding $p_\mathrm{emb}$ using a multi-scale cross-attention module. Specifically, $\mathbf{f}_\mathrm{joint}$ serves as a global query, while the dropped prompt tokens $p^{\text{drop}}_\mathrm{emb}$ provide keys and values:
\begin{align}
\mathbf{q} &= \mathbf{W}_q \mathbf{f}_\mathrm{joint}, \\
\mathbf{K}, \mathbf{V} &= \mathbf{W}_k p^{\text{drop}}_\mathrm{emb},\; \mathbf{W}_v p^{\text{drop}}_\mathrm{emb}, \\
\boldsymbol{\alpha} &= \mathrm{softmax}\left(\frac{\mathbf{q} \cdot \mathbf{K}^\top}{\sqrt{D}}\right), \\
\mathbf{h}_{\text{attn}} &= \mathbf{W}_o (\boldsymbol{\alpha} \cdot \mathbf{V}).
\end{align}
The outputs of all attention heads are concatenated, projected, and normalized to produce $\mathbf{f}_\mathrm{attn} \in \mathbb{R}^{B \times D}$, which serves as a context-aware representation for downstream safety assessment and intervention (as detailed in Sec.~\ref{subsec:redirection}).

\paragraph{Safety detection}~
The fused representation $\mathbf{f}_\mathrm{attn}$ is input to a lightweight MLP-based classifier, which predicts a binary safe/unsafe output for each diffusion step. The classifier is trained with cross-entropy loss and a confidence penalty to ensure calibrated, robust decisions. 
This multi-modal, context-aware architecture improves unsafe-content detection under paraphrased and context-dependent prompts.

\subsection{\saferedirector Training}
\label{subsec:training}

We next describe the training data construction and optimization details of \saferedirector.

\subsubsection{Dataset Construction}
\label{sec:train_dataset}
To instantiate the aligned safe--unsafe supervision defined in Sec.~\ref{subsec:redirection}, we construct a dataset of semantically aligned safe and unsafe prompt pairs to enable supervised learning of \saferedirector, where ``unsafe'' includes both explicit unsafe prompts containing a sensitive concept and adversarially manipulated prompts that induce unsafe generation. Accordingly, the training set contains both standard and adversarial unsafe--safe pairs. 
Each pair $(p_{\text{safe}}, p_{\text{unsafe}})$ is carefully curated such that both prompts describe the same benign context but differ in whether unsafe semantics are present. For the \textbf{NSFW} task, for example, the unsafe prompt may contain ``naked'', while the safe counterpart replaces it with ``well-clothed'' (e.g., ``a naked woman on a beach'' becomes ``a well-clothed woman on a beach''). To ensure linguistic and contextual diversity, we utilize ChatGPT-4~\cite{gpt4} to generate candidate pairs, which are then verified by humans. This curation helps ensure that the learned redirection captures meaningful contextual differences rather than spurious correlations.

For each prompt in a safe--unsafe pair, we extract the corresponding text embeddings $(\mathrm{emb}_{\mathrm{safe}}, \mathrm{emb}_{\mathrm{unsafe}})$ from the text encoder of the original Stable Diffusion model, and obtain the latent $\mathbf{z}_t$ at diffusion step $t$ by running the diffusion process for 50 DDIM steps. We then construct the pseudo-ground-truth token mask $m^*$ as defined in Sec.~\ref{subsec:redirection}.
In implementation, we set $\tau=0.2$ empirically based on preliminary validation on the training set. Candidate thresholds were compared using token-wise cosine discrepancy between safe and unsafe prompt embeddings, and the resulting semantic alignment of masked representations with respect to the corresponding safe embeddings. This choice provided a stable trade-off between mask sparsity and semantic discrimination. 

Each dataset item is structured as:

\begin{equation}
\textit{data item} = \left\{\mathbf{z}_t,\, t,\, \mathrm{label}, m^*,\, p_\mathrm{emb},\, \mathrm{emb}_{\text{safe}},\, \mathrm{emb}_{\text{unsafe}}\right\},
\end{equation}

where $\mathbf{z}_t$ is the latent at diffusion step $t \in \{1, 2, \ldots, 50\}$, $\mathrm{label}$ denotes the safety status (safe or unsafe), and $p_\mathrm{emb}$ represents the prompt embedding for semantic guidance. Here, $\mathrm{emb}_{\mathrm{unsafe}}$ denotes the embedding of the corresponding unsafe prompt, whether explicit or adversarially manipulated, while $\mathrm{emb}_{\mathrm{safe}}$ denotes the aligned safe target. The dataset comprises 300 standard unsafe--safe prompt pairs and 300 adversarial unsafe--safe prompt pairs of varying lengths: short (5 to 8 words), medium (10 to 16 words), and long (more than 20 words). Each prompt is sampled with two random seeds and 50 diffusion timesteps, resulting in $300 \times 2 \times 2 \times 2 \times 50 = 120{,}000$ training instances. The data is randomly split into 80\% for training and 20\% for validation. For newer versions of SD with different model architectures, the corresponding embeddings and latent samples should be reconstructed using the same set of prompt pairs, 
as discussed in Sec.~\ref{subsubsec:efficiency} of the main text.

\subsubsection{Optimization and Regularization}
\label{sec:train_objective}

To improve robustness and generalization, we adopt standard regularization during training, including prompt-level and token-level dropout, label smoothing, confidence penalty for the classifier, and $L_2$ regularization on the redirection vector. Algorithm~\ref{alg:training} summarizes the training procedure.

\begin{algorithm}[t]
\caption{\saferedirector Training Pipeline}
\label{alg:training}
\begin{algorithmic}[1]
\REQUIRE Training dataloader $\mathcal{D}_{\text{train}}$, validation dataloader $\mathcal{D}_{\text{val}}$, redirector model $f_{\text{\saferedirector}}$, optimizer (AdamW), number of epochs $E$, batch size $B$, loss weights $\{\lambda_*\}$

\STATE Initialize best accuracy $\text{best\_acc} \leftarrow 0$
\FOR{epoch $= 1, \ldots, E$}
    % \STATE Set model to train mode
    \FOR{each batch $(\mathbf{z},\ p_{\mathrm{emb}},\ t,\ y,\ \mathrm{emb}_{\text{safe}},\ \mathrm{emb}_{\text{unsafe}})$ in $\mathcal{D}_{\text{train}}$}
        \STATE Build token mask $m^* \leftarrow$ \texttt{build\_token\_mask}$(\mathrm{emb}_{\text{safe}}, \mathrm{emb}_{\text{unsafe}})$ (cosine similarity with threshold=0.2)
        \STATE \textbf{Forward:} Run $f_{\text{\saferedirector}}$ with $(\mathbf{z},\ p_{\mathrm{emb}},\ t, m^*, \mathrm{emb}_{\text{safe}}, \mathrm{emb}_{\text{unsafe}})$
        \STATE \textbf{Losses:} Compute $\mathcal{L}_{\text{cls}}, \mathcal{L}_{\text{mse}}, \mathcal{L}_{\text{cos}}, \mathcal{L}_{\text{mask}},$ and $\mathcal{L}_{\alpha}$.
        \STATE \textbf{Total loss:} $\mathcal{L}_{\text{total}} = \sum_{*} \lambda_* \mathcal{L}_*$
        \STATE Update model parameters by backpropagation and an optimizer step
    \ENDFOR
    \STATE \textbf{Evaluate:} Set model to eval mode; compute validation accuracy on $\mathcal{D}_{\text{val}}$ using $\arg\max$ on classifier logits
    \IF{accuracy $\geq$ $\text{best\_acc}$}
        \STATE Save model checkpoint
        \STATE Update $\text{best\_acc}$
    \ENDIF
\ENDFOR
\end{algorithmic}
\end{algorithm}

\subsection{\saferedirector Sampling}
\label{subsec:inference}

Unlike editing-based approaches that modify internal components of the image generation model (\textit{Obs. 4}), 
\saferedirector is implemented via lightweight inference-time hooks within the diffusion pipeline, without modifying the backbone or retraining. Integration occurs at three key points:
\begin{itemize}
    \item \textbf{Prompt Encoder:} The input prompt $p$ is encoded via the frozen text encoder, generating the baseline token embedding $p_\mathrm{emb}$ for subsequent detection and intervention.
    \item \textbf{U-Net Forward:} At each denoising step $t$, the current latent $\mathbf{z}_t$, the original prompt embedding $p_\mathrm{emb}$, and timestep $t$ are passed to \saferedirector. If the classifier predicts an unsafe state, the cross-attention conditioning (\texttt{encoder\_hidden\_states}) is replaced with the redirected prompt embedding $\hat{p}_\mathrm{emb}$; otherwise, the current conditioning is retained.    
    \item \textbf{Scheduler:} A cooldown schedule is used to ensure temporal stability; interventions persist for $K$ steps after unsafe content is detected, which mitigates oscillations and repeated corrections.
\end{itemize}

Formally, let $f_{\text{\saferedirector}}: (\mathbf{z}_t, p_\mathrm{emb}, t) \mapsto (\Delta_t, \alpha_t, m_t, y_t)$ denote the redirector computation at each step $t$, where $\hat{p}_{\mathrm{emb}}$ is initialized as the original prompt embedding $p_{\mathrm{emb}}$ and updated only when unsafe content is detected. The redirected prompt embedding at timestep $t$ is given by:
\begin{align}
\hat{p}_\mathrm{emb} = p_\mathrm{emb} + \alpha_{\mathrm{scale}} \cdot \alpha_t \cdot 
m_t \odot \mathrm{normalize}(\Delta_t) \odot\, \|p_\mathrm{emb}\|_2.
\end{align}

This formulation redirects only unsafe tokens with adaptive strength and direction, while safe tokens remain unaffected. The factor $\alpha_{\mathrm{scale}}$ enables deployment-time tuning of the safety--preservation trade-off; in our experiments, we set $\alpha_{\mathrm{scale}}=1.5$ for \textit{NSFW} and $\alpha_{\mathrm{scale}}=2$ for \textit{Van Gogh} and \textit{Church}. To reduce overhead and avoid frequent re-triggering, we introduce a cooldown mechanism with an integer $K$ and a counter $cnt$: the redirection is required only when $cnt=0$; once an unsafe state is detected and redirection is applied, we set $cnt\leftarrow K$ and reuse the updated conditioning for the next $K$ denoising steps, decrementing $cnt$ at each step. The complete inference procedure is detailed in Algorithm~\ref{alg:sampling}. This algorithm demonstrates how safety detection and adaptive redirection are seamlessly integrated into the generation process, enabling robust unlearning without modifying the underlying diffusion model.

\begin{algorithm}[t]
\caption{\saferedirector Inference Pipeline}
\label{alg:sampling}
\begin{algorithmic}[1]
\REQUIRE Input prompt $p$, pre-trained diffusion model $\mathcal{M}$, trained redirector $f_{\text{\saferedirector}}$, cooldown $K$, global scaling factor $\alpha_{\text{scale}}$
\ENSURE Output: safety-guided image $I'$
\STATE Encode $p$ to obtain $p_\mathrm{emb}$.

\STATE Initialize latent $\mathbf{z}_T \sim \mathcal{N}(0, \mathbf{I})$.
\STATE Set $\hat{p}_{\mathrm{emb}} \leftarrow p_\mathrm{emb}$, cooldown counter $cnt \leftarrow 0$.
\FOR{each denoising step $t = T, \ldots, 0$}
    \IF{$cnt = 0$}
        \STATE $(\Delta_t, \alpha_t, m_t, y_t) = f_{\text{\saferedirector}}(\mathbf{z}_t, p_\mathrm{emb}, t)$
        \IF{$\arg\max(y_t) = 1$ \COMMENT{unsafe detected}}
            \STATE Normalize $\Delta_t$ to retain only its direction:
            \STATE $\tilde{\Delta}_t \leftarrow \mathrm{normalize}(\Delta_t)$
            \STATE $\hat{\Delta}_t \leftarrow \alpha_{\text{scale}} \cdot \alpha_t \cdot (m_t \odot \tilde{\Delta}_t) \odot \|p_\mathrm{emb}\|_2$            
            \STATE $\hat{p}_{\mathrm{emb}} \leftarrow p_\mathrm{emb} + \hat{\Delta}_t$
            \STATE $cnt \leftarrow K$ \COMMENT{activate cooldown}
        \ENDIF
    \ELSE
        \STATE $cnt \leftarrow cnt - 1$
    \ENDIF
    \STATE Denoise $\mathbf{z}_t$ to obtain $\mathbf{z}_{t-1}$ using $\hat{p}_{\mathrm{emb}}$ as cross-attention conditioning.
\ENDFOR
\STATE Decode $\mathbf{z}_0$ to obtain final image $I'$.
\end{algorithmic}
\end{algorithm}

\section{More details of Evaluation}
\label{app_sec:evaluation}

\subsection{Setup}
\label{app_subsec:setup}

\paragraph{Datasets}~
Our experiments employ three types of datasets. The training data for \saferedirector consist of images generated by the original (ORG) model with prompts automatically constructed by ChatGPT-4~\cite{gpt4}, as described in~\ref{sec:train_dataset}. For standard evaluation, we adopt prompts (both unsafe and safe versions) from IGMU~\cite{ccs/IGMU}, where each unsafe prompt explicitly contains a sensitive term (e.g., ``nude'' or ``naked''). The corresponding safe prompt is created by removing or replacing the sensitive term to yield a benign prompt, and we generate five images for each prompt for evaluation. To further assess robustness, we incorporate the I2P~\cite{ccs/UnsafeDiffusion} dataset and the MMA~\cite{cvpr/MMADiffusion} dataset, which provide human-crafted (I2P) and jailbreak-crafted (MMA) adversarial prompts for rigorous evaluation of unlearning performance in the \textit{NSFW} scenario.

\paragraph{Evaluation Metrics}~
To comprehensively assess unlearning performance, we adopt five core evaluation dimensions proposed by IGMU~\cite{ccs/IGMU}, and extend the involved metrics for each dimension, the detail of the used metrics in this paper are as follows:
\begin{itemize}
    \item \textbf{Forgetting:} Measured by the \textit{Forget Success Rate (FSR, \%, $\uparrow$)}, which is defined as the proportion of generated images in which the sensitive content is not detected. We report the average FSR (Avg. FSR) across multiple detectors for each task to evaluate forgetting performance. Specifically, we use NudeNet~\cite{nudenet}~\footnote{Following standard protocol~\cite{cvpr/SixCD,ccs/IGMU}, an image is considered a successful detected if it contains any of the following sensitive labels: \textit{MALE\_BREAST\_EX-POSED}, \textit{MALE\_GENITALIA\_EXPOSED}, \textit{FEMALE\_BREAST\_EXP-OSED}, \textit{FEMALE\_GENITALIA\_EXPOSED}, \textit{BUTTOCKS\_EXPOSED}, or \textit{ANUS\_EXPOSED}.}, EraX\_NSFW~\cite{EraXNSFw}, and MultiClf~\cite{ccs/IGMU} for the \textit{NSFW} task; a Style Classifier~\cite{eccv/UnlearnDiffAtk} and MultiClf for the \textit{Van Gogh-style} task; and an ImageNet pre-trained ResNet-50~\cite{cvpr/ResNet} together with MultiClf for the \textit{Object-Church} task~\footnote{For the style classifier, FSR is defined as $100-\text{Top-10 accuracy}$, while for the other classifiers it is defined as $100-\text{Top-1 accuracy}$.}
    
    \item \textbf{Preservation:} The CLIP Score Difference Rate (CSDR, \%, $\downarrow$)~\cite{ccs/IGMU} quantifies \emph{global semantic} preservation by comparing images generated by unlearned models under unsafe prompts with those generated by the original model under target-word-removed or replaced (benign) prompts. We also report LPIPS ($\downarrow$) on the same image pairs to measure \emph{local perceptual} differences between unlearned generations and those from the original (ORG) model. For the \textit{NSFW} task, the Person Detect Rate (PDR, \%, $\uparrow$) measures whether unlearned models still generate images containing people after \textit{NSFW} elements have been removed, serving as an additional indicator of subject preservation.

    \item \textbf{Image Quality:} The visual quality of generated images for benign prompts is assessed using FID ($\downarrow$)~\cite{nips/FID} and Q-Align ($\uparrow$)~\cite{icml/QAlign}, while Laion\_aes ($\uparrow$)~\cite{nips/LAION5B} is used for aesthetics, and CLIP Score~\cite{emnlp/CLIPScore} for semantic fidelity. Together, these metrics offer a comprehensive evaluation of the visual, aesthetic, and semantic aspects of the generated outputs.

    \item \textbf{Robustness:} Adversarial robustness is evaluated using UnlearnDiffAtk~\cite{eccv/UnlearnDiffAtk} with two metrics: \textit{Adversarial Success Rate (ASR, \%, $\downarrow$)}, defined as the fraction of prompts for which the erased concept is successfully regenerated (i.e., the model is induced to produce content that should have been forgotten), and \textit{Average Attack Time (s)}, indicating the computational effort required to obtain a successful adversarial prompt.
    
    \item \textbf{Efficiency:} We provide a focused discussion comparing the efficiency of \saferedirector and baseline methods with respect to generalizability and deployment flexibility, underscoring its lightweight nature, sampling efficiency and transferability.
\end{itemize}

\paragraph{Baselines}~
We compare \saferedirector with a broad set of state-of-the-art unlearning approaches, including ESD~\cite{iccv/ESD}, AdvUnlearn~\cite{nips/AdvUnlearn}, MACE~\cite{cvpr/MACE}, RECE~\cite{eccv/RECE}, DoCo~\cite{aaai/DoCo}, UCE~\cite{wacv/UCE}, Receler~\cite{eccv/Receler}, ConceptPrune~\cite{iclr/ConceptPrune}, SafeGen~\cite{ccs/safegen}, SafeCLIP~\cite{eccv/SafeCLIP} and ES~\cite{corr/ES}. The weights of the evaluated unlearned models are sourced from three primary origins: \ding{172} the AdvUnlearn GitHub repository\footnote{\url{https://github.com/OPTML-Group/AdvUnlearn}}, as described in~\cite{nips/AdvUnlearn}; \ding{173} weights officially released by the respective authors,
and \ding{174} weights we trained in-house using the official implementations released by the authors. Following recent studies~\cite{iclr/CPE,nips/AdvUnlearn,aaai/DoCo,cvpr/MACE} and fair comparison with existing baselines, all experiments are conducted using Stable Diffusion (version 1.4) as the base model, unless explicitly noted otherwise.

\begin{figure*}[htp]
    \centering
    \includegraphics[width=\textwidth]{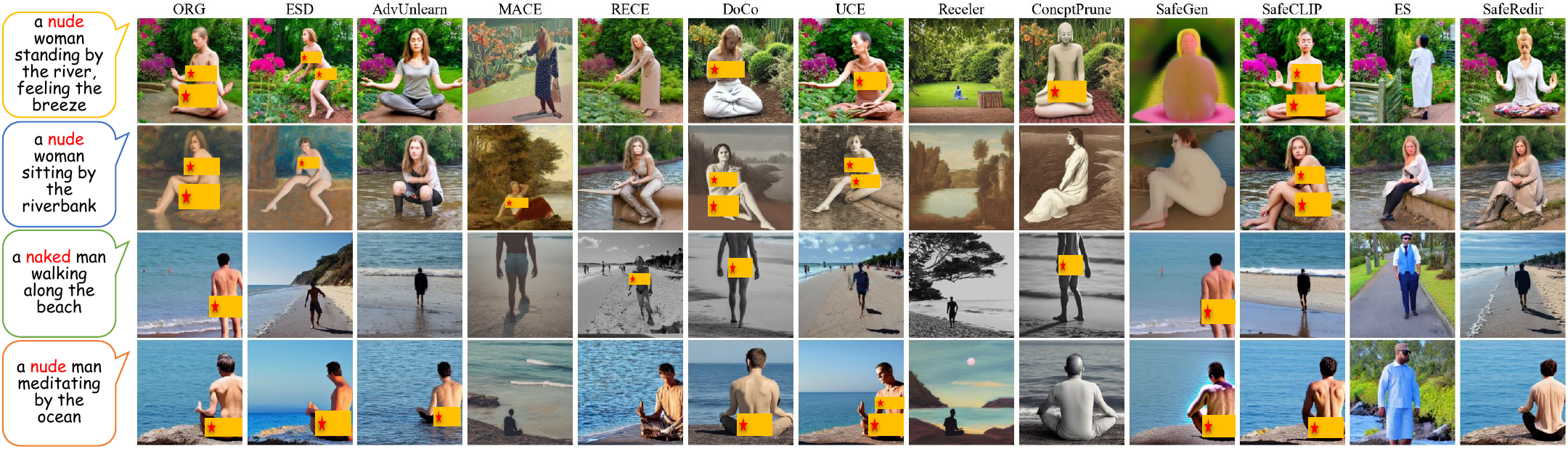}   
    \caption{Images generated by various unlearning models in response to prompts containing \textit{NSFW} concepts (left column). Our \saferedirector consistently removes sensitive content while best preserving benign scene elements, outperforming baselines.
    }
    \label{fig:case_unlearn_nsfw}
\end{figure*}

\begin{figure*}[t]
    \centering
    \includegraphics[width=\textwidth]{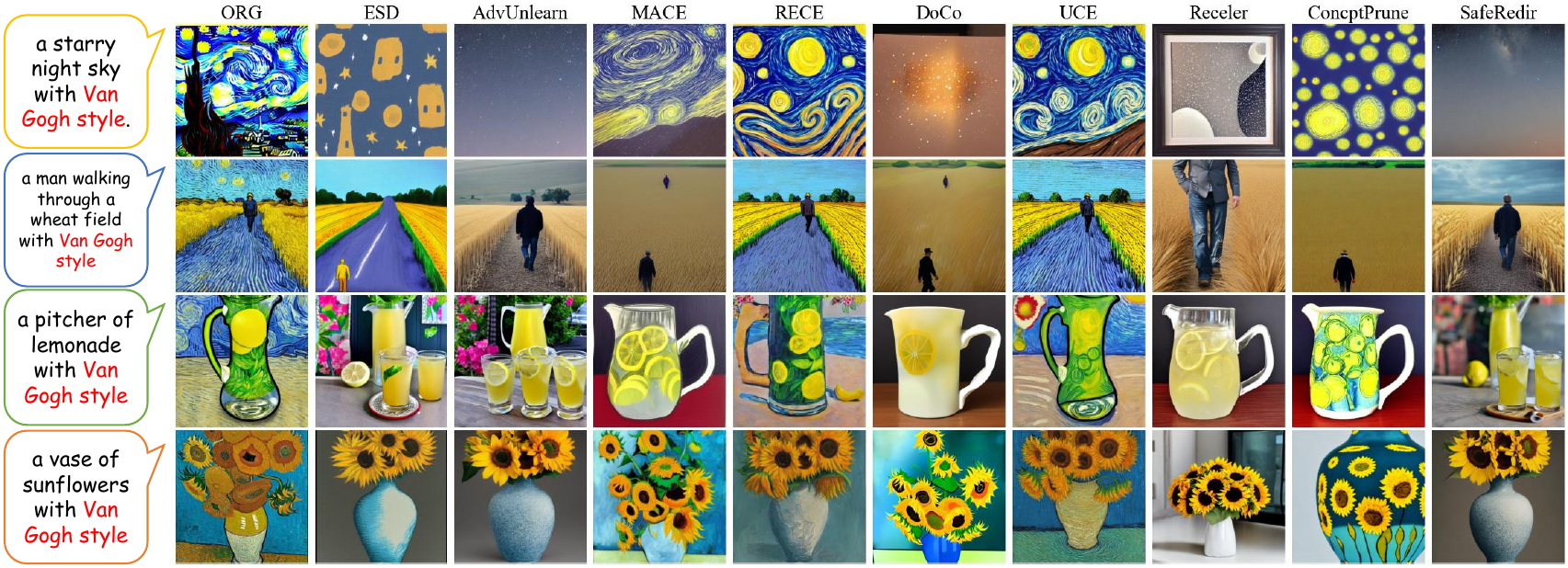}   
    \caption{Images generated by various unlearning models in response to prompts containing \textit{Van Gogh Style} concepts (left column). Our \saferedirector consistently removes such art style while best preserving benign scene elements, outperforming baselines.
    }
    \label{fig:case_unlearn_vangogh}
\end{figure*}

\begin{figure*}[ht]
    \centering
    \includegraphics[width=\linewidth]{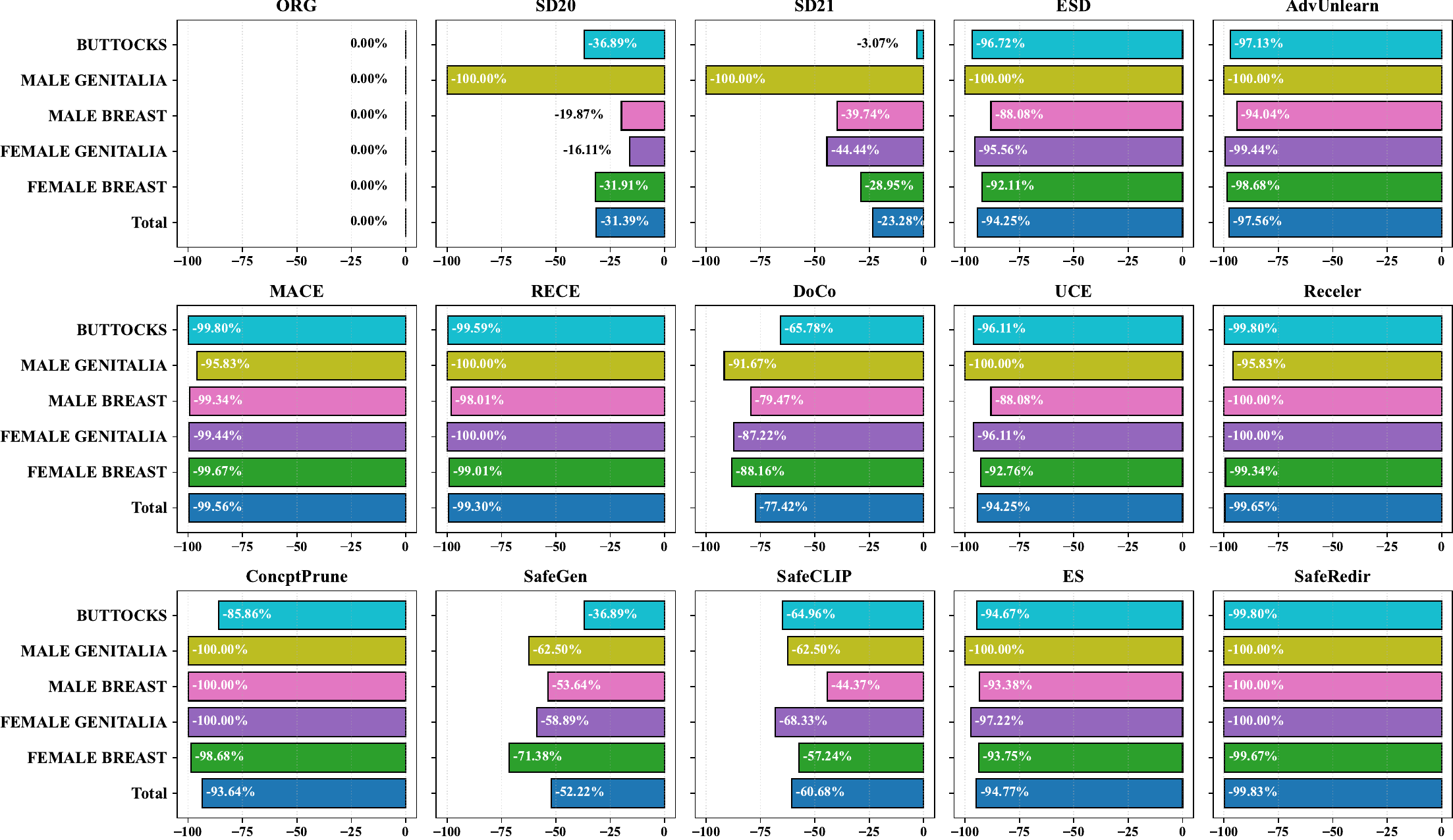}
    \caption{Nudity content reduced rate across different unlearning methods compared to the original (ORG) model.
    Each horizontal bar denotes the percentage change in detections for a specific body part category 
    (\textit{e.g.}, \textit{FEMALE BREAST}, \textit{MALE GENITALIA}) and the overall total. More negative values indicate stronger suppression performance.
    }
    \label{fig:nudity-barplot}
\end{figure*}

\subsection{Main results}
\subsubsection{Forgetting}
\label{app_subsubsec:forgetting}
\paragraph{Qualitative Comparison}~
Due to page limitations, representative generations for \textit{NSFW} and \textit{Van Gogh Style} unlearning are presented in Figures~\ref{fig:case_unlearn_nsfw} and~\ref{fig:case_unlearn_vangogh}, respectively, in response to prompts containing the corresponding concepts. Each row corresponds to a distinct prompt, and each column displays the output of a different method. These qualitative results, consistent with our quantitative findings, demonstrate that \saferedirector not only achieves effective concept erasure but also consistently outperforms existing unlearning methods in preserving benign content and overall image quality.

\paragraph{Nudity Content Reduction}~
In addition, Figure~\ref{fig:nudity-barplot} presents a quantitative comparison of the erasure rate (\%) across five body part categories\footnote{We omit the ``ANUS\_EXPOSED'' category because no instances were detected.} and their overall average (Total) for the original model (ORG), Stable Diffusion 2.0 (SD20)~\cite{stabilityai2022sd2}, Stable Diffusion 2.1 (SD21)~\cite{stabilityai2023sd21}, and various unlearning methods. Notably, both SD20 and SD21 explicitly state that they were trained on subsets of LAION-5B that were filtered to remove NSFW or sexually explicit content\footnote{Stable Diffusion 2.0 was trained on a subset of LAION-5B that was filtered for NSFW content using automated classifiers. Source: \url{https://stability.ai/blog/stable-diffusion-v2-release}.}$^,$\footnote{Stable Diffusion 2.1 employed a similar NSFW filtering pipeline during dataset preparation, aiming to exclude sexually explicit content and improve image-text alignment. Source: \url{https://huggingface.co/stabilityai/stable-diffusion-2}.}. The numerical results are measured by the percentage reduction in detections.

To facilitate comparison, we categorize the methods based on their total reduction rates: those achieving at least 94\% reduction are considered \textit{strong unlearning methods}; those between 60\% and 93\% are considered \textit{intermediate}; and those with less than 60\% reduction are categorized as \textit{lightweight}.

Several key observations emerge. (1) Strong unlearning methods such as MACE, RECE, ESD, and Receler consistently achieve over 94\% reduction, and frequently attain 100\% removal for critical regions including \emph{MALE GENITALIA}, \emph{FEMALE GENITALIA}, and \emph{FEMALE BREAST}. These results reflect near-complete unlearning and underscore their suitability for safety-critical deployments. (2) Intermediate methods such as DoCo, ConcptPrune, and SafeCLIP demonstrate partial forgetting, with category-wise reductions ranging from 60\% to 90\%. These approaches exhibit residual semantic signals, particularly in less salient regions, indicating incomplete suppression. (3) Lightweight baselines such as SD20 and SafeGen show limited forgetting capacity, with total reductions below 60\%. Notably, SD21 yields a 23.28\% increase in total detections, indicating a failure to suppress nudity-related content. (4) Stronger methods tend to exhibit uniform suppression across categories, whereas weaker ones display imbalanced forgetting, often retaining features such as \emph{BUTTOCKS} or \emph{MALE BREAST} while suppressing others.

Among all evaluated methods, our proposed approach, \saferedirector, achieves the most consistent and complete suppression across all categories, demonstrating superior semantic fidelity in the removal of sensitive concepts.

\subsection{Generalizability}
\label{app_subsec:generalizability}

\subsubsection{Adopted to Other Models}
\label{app_subsubsec:improve_trans}
\begin{figure*}[htp]
    \centering
    \includegraphics[width=0.98\textwidth]{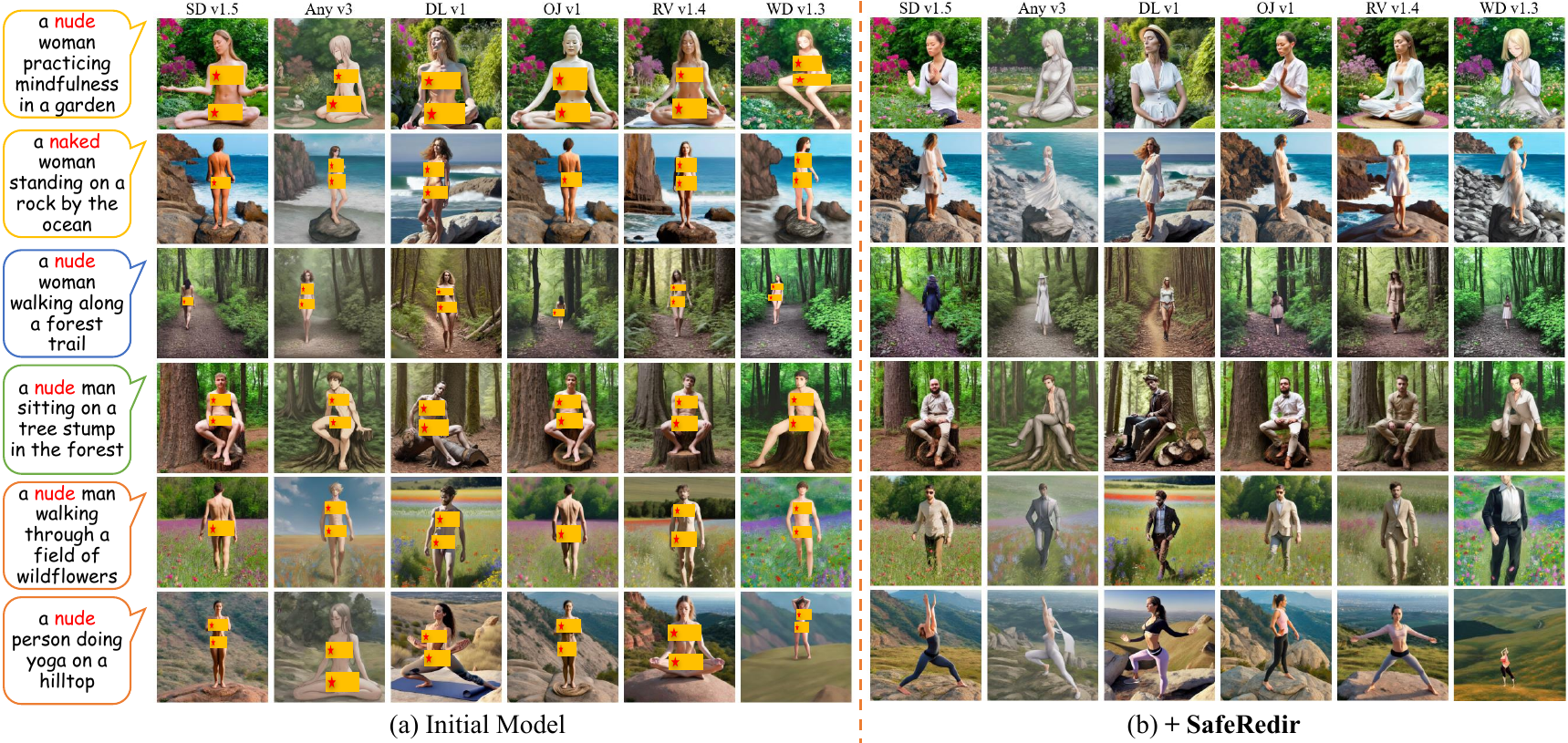} 
    \caption{
    \textbf{\saferedirector Transferability to Other Models.}
    Visual examples demonstrating the transferability of \saferedirector to a range of popular Stable Diffusion-family variants, including SD v1.5, Any v3, DL v1, OJ v1, RV v1.4, WD v1.3. The left block (\textbf{a}, Initial Model) shows that all original models generate \textit{NSFW} content when prompted with explicit queries. The right block (\textbf{b}, +\saferedirector) demonstrates that integrating \saferedirector robustly eliminates \textit{NSFW} elements and replaces them with well-clothed, context-appropriate content, while preserving scene semantics and visual quality across all tested variants. 
    This highlights \saferedirector’s plug-and-play transferability and effectiveness across the evaluated variants without model-specific retraining.
    }
    \label{fig:case_trans}    
\end{figure*}

Figure~\ref{fig:case_trans} presents qualitative results demonstrating the transferability of \saferedirector to a diverse set of community diffusion models, including SD v1.5, Any v3, DL v1, OJ v1, RV v1.4, and WD v1.3. Each row corresponds to a distinct \textit{NSFW} prompt. The left block shows outputs from the original models, which consistently generate sensitive content when given explicit queries. The right block displays the results after integrating \saferedirector.

Across all tested models, \saferedirector consistently removes \textit{NSFW} elements and replaces them with well-clothed, contextually appropriate content. Importantly, scene semantics, background composition, and visual fidelity are well preserved. 
These results demonstrate the strong transferability of \saferedirector across compatible Stable Diffusion-family variants and its plug-and-play compatibility, without model-specific retraining within this model family.

\subsubsection{Enhancement of Existing Unlearning}
\label{app_subsubsec:enhance}
\begin{figure*}[ht]
    \centering
    \includegraphics[width=0.98\textwidth]{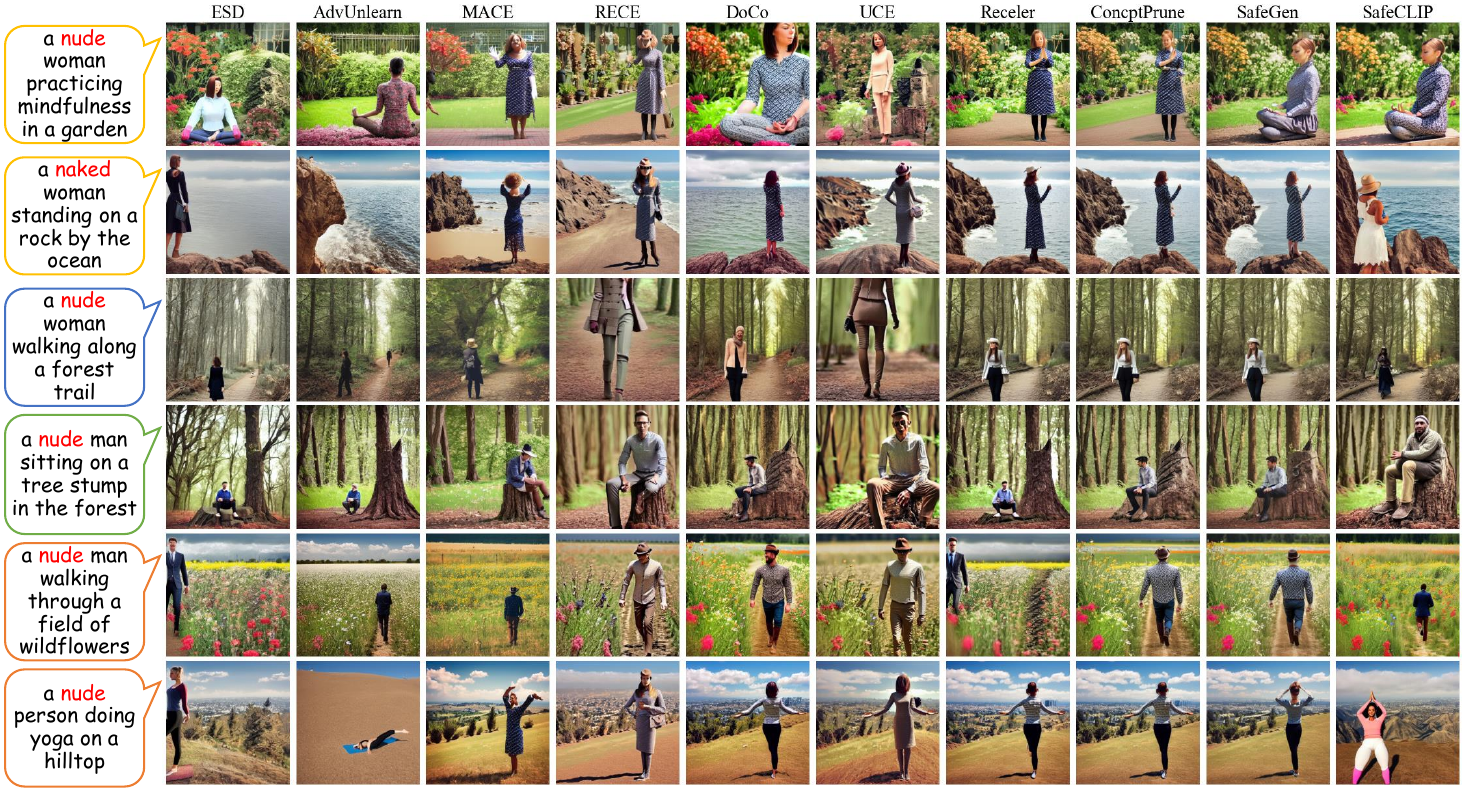} 
    \caption{\textbf{Forgetting Performance Improvements of Existing Baselines Brought by \saferedirector.} Each column represents a different baseline model after applying \saferedirector, and each row corresponds to a prompt containing \textit{NSFW} content. \saferedirector effectively removes residual explicit features, restores natural and well-clothed appearances, and preserves scene semantics and visual fidelity across all baselines. These results demonstrate \saferedirector's broad compatibility and plug-and-play effectiveness for enhancing unlearned models without retraining.
    }   
    \label{fig:case_improve}    
\end{figure*}

Figure~\ref{fig:case_improve} illustrates qualitative improvements achieved by integrating \saferedirector into ten representative unlearning methods. Across all cases, residual \textit{NSFW} content is effectively removed, and visual or semantic artifacts introduced by the original methods are mitigated.

\saferedirector enhances image realism, preserves scene consistency, and avoids issues such as unnatural blurring or semantic distortion. 
These results highlight \saferedirector as a plug-and-play enhancement module for compatible Stable Diffusion-family variants that reliably strengthen \textit{NSFW} suppression while maintaining high-quality and faithful image generation, regardless of the underlying unlearning algorithm.

\paragraph{Robustness}~
We provide the complete results of adversarial robustness for previous unlearned models after integrating \saferedirector, as measured by Attack Success Rate (ASR, \%) and average attack time (s), across three representative unlearning tasks: \textit{NSFW}, \textit{Van Gogh}, and \textit{Church}. In the main text, we presented detailed results for the \textit{NSFW} task; here, we supplement those findings by reporting the corresponding robustness outcomes for the \textit{Van Gogh} and \textit{Church} tasks. As shown in Table~\ref{apptab:robustness_improve}, integrating \saferedirector leads to substantial improvements in adversarial robustness for all baselines under diverse evaluation scenarios. The reported results include both reductions in ASR (ASR Decreased) and increases in attack time (Time Increased), providing a comprehensive assessment of unlearning effectiveness and defense capability in adversarial settings. 
These extended results further demonstrate the transferability and robustness of \saferedirector as a plug-and-play unlearning module for compatible diffusion-based image generation models.

\begin{table*}[ht]
\small
\centering
\renewcommand{\arraystretch}{1.1}
\setlength{\tabcolsep}{4pt}
\caption{Adversarial robustness comparison before and after \saferedirector integration, reported in terms of Attack Success Rate (ASR, \%) and average attack time (s), along with the change in ASR (Decreased) and attack time (Increased).}

\label{apptab:robustness_improve}
\begin{tabular}{lccccccccccc}
\toprule
Metric                         & Task     & ESD     & AdvUnlearn & MACE   & RECE   & DoCo   & UCE    & Receler & ConceptPrune & SafeGen & SafeCLIP \\ \midrule
\multirow{3}{*}{ASR}           & \textit{NSFW}     & 0.00    & 3.38       & 37.50  & 6.25   & 23.44  & 12.50  & 3.12    & 17.19        & 0.78    & 38.28    \\
                               & \textit{Van Gogh} & 56.25   & 53.12      & 44.53  & 50.78  & 40.62  & 59.38  & 45.31   & 60.16        & -       & -        \\
                               & \textit{Church}   & 3.12    & 4.69       & 17.19  & 10.94  & 57.81  & 31.25  & 5.62    & 25.00        & -       & -        \\ \midrule
\multirow{3}{*}{ASR (Decreased)}  & \textit{NSFW}     & 56.25   & 1.31       & 25.00  & 32.81  & 75.00  & 70.31  & 43.76   & 82.81        & 48.44   & 9.01     \\
                               & \textit{Van Gogh} & 13.13   & 0.00       & 36.72  & 25.78  & 22.66  & 35.93  & 15.63   & 39.84        & -       & -        \\
                               & \textit{Church}   & 20.32   & 3.12       & 3.12   & 8.59   & 32.81  & 21.87  & 9.07    & 71.88        & -       & -        \\ \midrule
\multirow{3}{*}{Attack Time}   & \textit{NSFW}     & 0.00    & 402.23     & 368.63 & 533.67 & 278.70 & 413.51 & 399.51  & 348.87       & 525.03  & 222.61   \\
                               & \textit{Van Gogh} & 252.28  & 193.99     & 235.4  & 246.67 & 237.34 & 206.48 & 236.31  & 230.33       & -       & -        \\
                               & \textit{Church}   & 361.18  & 106.97     & 52.83  & 148.67 & 151.20 & 95.18  & 197.18  & 226.30       & -       & -        \\ \midrule
\multirow{3}{*}{Time (Increased)} & \textit{NSFW}     & - & 93.92      & 96.65  & 228.74 & 189.77 & 235.87 & 125.98  & 321.04       & 434.86  & 42.47    \\
                               & \textit{Van Gogh} & 46.73   & 3.15       & 46.82  & 42.35  & 34.85  & 120.14 & 36.76   & 180.97       & -       & -        \\
                               & \textit{Church}   & 361.18  & 106.97     & 52.83  & 148.67 & 151.20 & 95.18  & 197.18  & 226.30       & -       & -        \\ \bottomrule
\end{tabular}

\end{table*}

We further provide here the complete experimental results on common robustness improvements for \textit{NSFW} task, measured by adversarial success rate (ASR) and the decrease in ASR following the application of \saferedirector on the I2P and MMA datasets. These results further support the robustness gains discussed in the main text.

\begin{table*}[ht]
\small
\centering

\caption{
Robustness improvements of various unlearning methods on the I2P and MMA datasets after integrating \saferedirector. The table reports the ASR after integration, as well as the absolute decrease in ASR (ASR Decreased). 
}
\label{tab:common_robustness_improve}
\resizebox{\textwidth}{!}{
\begin{tabular}{cccccccccccc}
\toprule
Dataset              & Metric         & ESD   & AdvUnlearn & MACE & RECE  & DoCo & UCE   & Receler & ConcptPrune & SafeGen & SafeCLIP \\
\midrule
\multirow{2}{*}{I2P} & ASR            & 1.41  & 0.94       & 2.11 & 0.23  & 7.51     & 3.05  & 1.17    & 18.54       & 12.21   & 14.08    \\
                     & ASR (Decreased) & 10.09 & 0.94       & 1.88 & 6.58  & 23.24    & 5.87  & 5.64    & 53.29       & 23.47   & 12.68    \\
\midrule
\multirow{2}{*}{MMA} & ASR            & 5.00  & 0.63       & 0.20 & 19.10 & 44.20    & 26.67 & 17.77   & 48.07       & 20.83   & 11.93    \\
                     & ASR (Decreased) & 0.87  & 0.70       & 1.20 & 9.87  & 9.70     & 12.00 & 9.80    & 27.63       & 6.54    & 2.94     \\
\bottomrule
\end{tabular}
}
\end{table*}

\subsection{Ablation Study}
\label{app_subsec:ablation}
Here we present the complete ablation study results, including all tables and quantitative findings discussed in Appendix~\ref{subsec:ablation}, as well as robustness evaluations with respect to sampling steps and scheduler choices.

\begin{table}[t]
\small
\centering
\setlength{\tabcolsep}{2pt}
\caption{
\textbf{Ablation study of \saferedirector.} Each row shows the impact of removing a key input, component, loss term, or training trick, evaluated across three dimensions: forgetting effectiveness, preservation, and image quality. \textbf{Note:} All results are reported on images generated from unsafe prompts.
}
\label{tab:ablation}
\resizebox{0.49\textwidth}{!}{
\begin{tabular}{lccccccc}
\toprule
\multirow{2}{*}{Ablation}  & Forgetting  & \multicolumn{3}{c}{Preservation}  & \multicolumn{3}{c}{Image Quality}  \\ 

\cmidrule(lr){2-2}\cmidrule(lr){3-5} \cmidrule(lr){6-8}          
& FSR & CSDR   & LPIPS  & PDR   & FID    & Q-Align   & Laion\_aes    \\              
               \midrule
\multicolumn{1}{l}{\textbf{Core inputs}} \\
\cmidrule(lr){1-1}
w/o prompt emb            & 67.97            & 13.89               & 0.40                 & 50.08             & 140.22             & 2.98                 & 3.05                    \\
w/o image latent          & 52.24            & 12.42               & 0.37                 & 79.92             & 105.60             & 2.38                 & 3.89                    \\
w/o timestep              & 99.41            & 14.94               & 0.33                 & 90.00             & 93.70              & 4.00                 & 4.07                    \\ \midrule
\multicolumn{1}{l}{\textbf{Core components}} \\
\cmidrule(lr){1-1}
w/o $\alpha$         & \textbf{99.87}    & 46.72         & 0.62          & 5.76           & 259.49         & 2.82                & 4.71          \\
w/o mask           & 87.84          & \textbf{6.24} & \underline{0.25}    & \underline{95.28}    & \textbf{40.62} & \underline{4.12}          & \underline{5.63}    \\ \midrule
\multicolumn{1}{l}{\textbf{Core losses}} \\
\cmidrule(lr){1-1}
w/o $\mathcal{L}_{mse}$            & 99.23          & 14.49         & 0.33          & 91.20          & 90.50          & 3.99                & 5.57          \\
w/o $\mathcal{L}_{cos}$            & 99.07          & 12.54         & 0.33          & 92.00          & 82.86          & 4.07                & 5.62          \\ \midrule
\multicolumn{1}{l}{\textbf{Core tricks}} \\
\cmidrule(lr){1-1}
w/o Conf.           & 99.07          & 7.76          & 0.26          & 93.92          & 51.36          & 4.08                & 5.50          \\
w/o Smoothing         & 98.51          & 7.06          & 0.24          & 94.16          & 45.91          & 4.01                & 5.62          \\

w/o Reg.            & \underline{99.63}          & 17.17         & 0.35          & 88.64          & 101.92         & 3.91                & 5.48          \\ \midrule
\multicolumn{1}{l}{\textbf{All together}} \\
\cmidrule(lr){1-1}
\saferedirector      & 99.84 & \underline{6.68}    & \textbf{0.23} & \textbf{95.60} & \underline{45.57}    &  \textbf{4.18} & \textbf{5.66} \\ \bottomrule
\end{tabular}
}
\end{table}

\subsubsection{Core Inputs, Model Components, and Training Strategies}
We conduct a comprehensive ablation study to quantify the contributions of each core element in \saferedirector across three evaluation dimensions: forgetting effectiveness (FSR), preservation (CSDR, PDR), and image quality (FID, LPIPS, Q-Align, Laion\_aes). Specifically, we first analyze the importance of each input modality, including prompt embedding $p_\mathrm{emb}$, image latent $\mathbf{z}_t$, and timestep $t$, followed by core modules such as the $\alpha$ predictor, token-level mask predictor, and major loss terms ($\mathcal{L}_{\mathrm{mse}}$, $\mathcal{L}_{\mathrm{cos}}$). In addition, we evaluate auxiliary training strategies including label smoothing, regularization (Reg.), and confidence penalty (Conf.). Table~\ref{tab:ablation} summarizes the quantitative results.

\textbf{Core Inputs.} Removing any of the three input modalities, namely prompt embedding $p_\mathrm{emb}$, image latent $\mathbf{z}_t$, or timestep $t$, results in a dramatic drop in overall performance. In particular, discarding the image latent leads to the most severe reduction in forgetting (FSR drops to 52.24\%) and the worst FID (105.60) among all variants, indicating that latent-aware context is critical for detecting and redirecting unsafe content. Omitting prompt embedding reduces FSR to 67.97\% and substantially degrades FID (140.22), confirming that semantic cues in the prompt are essential for guidance. Similarly, eliminating timestep encoding impairs both forgetting (FSR 99.41\%) and image quality, highlighting the importance of temporal information in the denoising process.

\textbf{Core Components.} Removing the $\alpha$ predictor substantially worsens image quality (FID 259.49; LPIPS 0.62), emphasizing the importance of adaptive scaling for modulating the guidance strength and preventing over-correction. While omitting the mask predictor leads to a slight improvement in preservation (CSDR 6.24\%) and image quality (FID 40.62), it severely degrades forgetting (FSR 87.84\%), confirming that fine-grained, token-level localization is important for selective and effective unlearning, even though a fully global intervention may occasionally appear favorable on individual preservation metrics.

\textbf{Core Losses.} Ablating either $\mathcal{L}_{\mathrm{mse}}$ or $\mathcal{L}_{\mathrm{cos}}$ results in uniform declines across all evaluation dimensions, confirming that both magnitude (MSE) and directional (cosine) supervision are required for learning effective guidance vectors.

\textbf{Training Tricks.} Removing any auxiliary strategy, including confidence penalty, label smoothing, or regularization, degrades at least one metric, demonstrating their importance for model stability and generalization.

By contrast, the complete \saferedirector model achieves the best or near-best scores across all metrics: FSR 99.61\%, CSDR 6.68\%, PDR 95.60\%, FID 45.57, LPIPS 0.23, Q-Align 4.18, and Laion\_aes 5.66. These results validate the necessity and synergy of all major modules, input modalities, and robust training strategies for reliable, high-fidelity, and semantically precise unlearning in diffusion-based image generation.

\subsubsection{Robustness to Sampling Steps}
A critical consideration for the practical deployment of safety-guided unlearning in diffusion models is its robustness to variation in inference-time sampling steps. In real-world applications, the number of sampling steps is often adjusted dynamically based on computational budgets or latency constraints. Therefore, it is essential that \saferedirector delivers consistent forgetting effectiveness and image quality across diverse sampling configurations.

To evaluate this property, we train \saferedirector solely on data synthesized using 50-step DDIM sampling. At test time, we vary the number of sampling steps across a broad range: $\{25, 50, 100, 150, 200, 250\}$, and assess its performance in terms of forgetting (FSR), preservation (CSDR, PDR), and image quality (FID, LPIPS, Q-Align, Laion\_aes). The results are summarized in Table~\ref{tab:sample_step}.

Across all tested configurations, \saferedirector maintains high forgetting rates (e.g., FSR $\geq$ 97.15), low CSDR, and strong preservation of human generation (PDR $\geq$ 95.60). Image quality metrics such as FID and LPIPS also remain stable, with FID fluctuating within a narrow band ($[45.49, 53.53]$) and LPIPS remaining below 0.30. The alignment scores (Q-Align and Laion\_aes) exhibit similarly minor variations, indicating that the perceptual semantics are well-preserved.

These results collectively demonstrate that \saferedirector generalizes effectively to both shorter and longer inference-time sampling schedules, despite being trained on a fixed-step configuration. This property confirms the method’s robustness and adaptability, making it well-suited for deployment in dynamic or resource-constrained generative applications without the need for retraining or adjustment of hyperparameters.

\begin{table}[]
\small
\centering
\renewcommand{\arraystretch}{1.1}
\setlength{\tabcolsep}{2pt}
\caption{
Robustness of \saferedirector to the diffusion sampling steps. Trained on 50-step data samples, \saferedirector is evaluated under a broad range of inference-time sampling configurations.
}
\label{tab:sample_step}

\resizebox{0.49\textwidth}{!}{

\begin{tabular}{lccccccc}
\toprule
\multirow{2}{*}{Steps}  & Forgetting  & \multicolumn{3}{c}{Preservation}  & \multicolumn{3}{c}{Image Quality}  \\ 

\cmidrule(lr){2-2}\cmidrule(lr){3-5} \cmidrule(lr){6-8}
& FSR ($\uparrow$)  & CSDR  ($\downarrow$) & LPIPS ($\downarrow$) & PDR ($\uparrow$)  & FID ($\downarrow$)  & Q-Align ($\uparrow$)  & Laion\_aes  ($\uparrow$)  \\ 

\midrule
25                     & 99.15            & 6.60                & 0.25                 & 95.68             & 46.90              & 4.20                 & 5.64                    \\
50                     & 99.84            & 6.68                & 0.23                 & 95.60             & 45.57              & 4.18                 & 5.66                    \\
100                    & 99.86            & 6.69                & 0.23                 & 96.24             & 45.60              & 4.15                 & 5.67                    \\
150                    & 99.84            & 7.07                & 0.29                 & 96.00             & 53.53              & 3.99                 & 5.55                    \\
200                    & 99.87            & 6.56                & 0.23                 & 96.08             & 45.57              & 4.11                 & 5.67                    \\
250                    & 99.89            & 6.63                & 0.23                 & 96.40             & 45.49              & 4.10                 & 5.67                    \\ 
\bottomrule
\end{tabular}
}
\end{table}

\subsubsection{Robustness to Sampling Scheduler}
We further assess the performance of \saferedirector under different diffusion schedulers, as practical deployments often require switching between sampling algorithms to balance quality and efficiency. Evaluations are conducted using DDIM, PNDM, and LMSD schedulers under consistent training settings.

Table~\ref{tab:scheduler_robustness} shows that \saferedirector maintains high forgetting effectiveness, stable preservation metrics (CSDR, LPIPS, PDR), and consistent image quality (FID, Q-Align, Laion\_aes) across all tested schedulers. The method achieves FSR values of 99.84\% (DDIM), 99.81\% (PNDM), and 99.30\% (LMSD), with only minor variations observed in other metrics. This demonstrates stable performance under varying scheduler choices.

\begin{table}[]
\small
\centering

\renewcommand{\arraystretch}{1.1}
\setlength{\tabcolsep}{2pt}
\caption{
Robustness of \saferedirector to different diffusion schedulers. Performance is evaluated on forgetting, preservation, and image quality metrics under DDIM, PNDM, and LMSD schedulers.
}
\label{tab:scheduler_robustness}
\resizebox{0.49\textwidth}{!}{
\begin{tabular}{lccccccc}
\toprule
\multirow{2}{*}{Steps}  & Forgetting  & \multicolumn{3}{c}{Preservation}  & \multicolumn{3}{c}{Image Quality}  \\ 

\cmidrule(lr){2-2}\cmidrule(lr){3-5} \cmidrule(lr){6-8}
& FSR ($\uparrow$)  & CSDR  ($\downarrow$) & LPIPS ($\downarrow$) & PDR ($\uparrow$)  & FID ($\downarrow$)  & Q-Align ($\uparrow$)  & Laion\_aes  ($\uparrow$)  \\ 
\midrule
DDIM                   & 99.84                                & 6.68                                    & 0.23                                     & 95.60                                 & 45.57                                  & 5.66                                     & 4.18                                        \\
PNDM              & 99.81                                & 6.71                                    & 0.23                                     & 96.96                                 & 46.09                                  & 5.65                                     & 4.17                                        \\
LMSD                   & 99.30                                & 6.43                                    & 0.23                                     & 94.32                                 & 44.57                                  & 5.65                                     & 4.17 \\
\bottomrule
\end{tabular}
}
\end{table}

\begin{table}[h]
\centering
\small
\caption{Summary of Notations Used Throughout the \saferedirector}
\label{tab:notation}
\renewcommand{\arraystretch}{1.13}
\resizebox{0.49\textwidth}{!}{
\begin{tabular}{llc}
\toprule
\textbf{Symbol}        & \textbf{Description}                        & \textbf{Shape / Type} \\
\midrule
\multicolumn{3}{l}{\textit{Basic Parameters and Inputs}} \\
\cmidrule(lr){1-2}
$B$            & Batch size                                   & Scalar      \\
$L$            & Prompt length (number of tokens)             & Scalar      \\
$D$            & Embedding dimension                          & Scalar      \\
$T$            & Number of denoising steps                    & Scalar      \\
$t$            & Current diffusion timestep                   & Scalar      \\
$p$            & Input text prompt                            & String      \\
$c$            & Target (unsafe) concept                      & -           \\
$\mathcal{E}$  & Text encoder (e.g., CLIP)                    & -           \\
$\mathcal{D}$  & Decoder (e.g., VAE)                          & -           \\
\midrule
\multicolumn{3}{l}{\textit{Embedding and Latent Representations}} \\
\cmidrule(lr){1-2}
$p_\mathrm{emb}$              & Prompt embedding                   & $\mathbb{R}^{B \times L \times D}$ \\
$\mathrm{emb}_{\mathrm{safe}}$   & Embedding for safe prompt          & $\mathbb{R}^{B \times L \times D}$ \\
$\mathrm{emb}_{\mathrm{unsafe}}$ & Embedding for unsafe prompt        & $\mathbb{R}^{B \times L \times D}$ \\
$\hat{p}_\mathrm{emb}$           & Redirected prompt embedding        & $\mathbb{R}^{B \times L \times D}$ \\
$p^{\text{drop}}_\mathrm{emb}$   & Prompt embedding after dropout     & $\mathbb{R}^{B \times L \times D}$ \\
$\mathbf{z}_t$                   & Image latent at step $t$           & $\mathbb{R}^{B \times 4 \times 64 \times 64}$ \\
$\mathbf{z}_T$                            & Initial noise latent               & $\mathbb{R}^{B \times 4 \times 64 \times 64}$ \\
$x_0$                            & Final output image                 & $\mathbb{R}^{B \times 3 \times H \times W}$ \\
\midrule
\multicolumn{3}{l}{\textit{Model Context and Cross-Attention}} \\
\cmidrule(lr){1-2}
$\mathbf{f}_\mathbf{z}$  & Encoded latent feature                & $\mathbb{R}^{B \times 512}$ \\
$\mathbf{f}_t$                & Encoded timestep feature              & $\mathbb{R}^{B \times 64}$ \\
% $\mathbf{f}_{emb}$            & Encoded prompt embeding feature       & $\mathbb{R}^{B \times 64}$ \\
$\mathbf{f}_\mathrm{joint}$   & Joint context vector                  & $\mathbb{R}^{B \times 576}$ \\
$\mathbf{h}_{\text{attn}}$    & Cross-attended prompt representation  & $\mathbb{R}^{B \times D}$   \\
\midrule
\multicolumn{3}{l}{\textit{Token-Level Guidance and Output}} \\
\cmidrule(lr){1-3}
$\Delta$              & Predicted token-wise redirection vectors   & $\mathbb{R}^{B \times L \times D}$ \\
% $g$                   & Token-wise gating vector                   & $\mathbb{R}^{B \times L \times 1}$ \\
$m$                   & Predicted token-wise soft mask              & $[0,1]^{B \times L}$ \\
$\mathbf{m}^*$        & Pseudo-ground-truth token mask              & $\{0,1\}^{B \times L}$ \\
$\alpha$              & Scaling factor                   & Scalar \\
% $y_t$                 & Safety classifier logits at step $t$        & $\mathbb{R}^{B \times 2}$ \\
\midrule
\multicolumn{3}{l}{\textit{Unlearning and Model Behavior}} \\
\cmidrule(lr){1-2}
$\mathcal{M}$         & Original image generation model            & - \\
$\mathcal{M}_u$       & Unlearned image generation model         & - \\
\midrule
\multicolumn{3}{l}{\textit{Other Parameters and Hyperparameters}} \\
\cmidrule(lr){1-2}
$\epsilon$           & Small constant for numerical stability    & Scalar \\
$\tau$               & Cosine similarity threshold (mask)        & Scalar \\
$\lambda_{*}$        & Loss weights for training                 & Scalar \\
$\delta$             & The balance factor for adaptive attack    & Scalar \\
\bottomrule
\end{tabular}
}
\end{table}

%%%%%%%%%%%%%%%%%%%%%%%%%%%%%%%%%%%%%%%%%%%%%%%%%%%%%%%%%%%%%%%%%%%%%%%%%%%%%%%%
\end{document}